\pdfoutput=1

\documentclass[11pt]{article}

\usepackage[preprint]{acl}

\usepackage{times}
\usepackage{latexsym}

\usepackage[T1]{fontenc}

\usepackage[utf8]{inputenc}

\usepackage{microtype}

\usepackage{inconsolata}

\usepackage{graphicx}

\usepackage{booktabs}
\usepackage{multirow}
\usepackage{subcaption}
\usepackage{rotating}
\usepackage{comment}
\usepackage{makecell}
\usepackage{soul}
\usepackage{xcolor}
\usepackage{xpatch}
\usepackage{enumitem}
\usepackage{CJKutf8}
\newcommand{\zh}[1]{\protect\begin{CJK*}{UTF8}{gbsn}#1\end{CJK*}}
\usepackage{devanagari}

\usepackage{amsmath,amsfonts,bm}
\usepackage{xspace}

\def\eqref#1{equation~\ref{#1}}

\def\1{\bm{1}}

\DeclareMathAlphabet{\mathsfit}{\encodingdefault}{\sfdefault}{m}{sl}
\SetMathAlphabet{\mathsfit}{bold}{\encodingdefault}{\sfdefault}{bx}{n}

\newcommand{\Loss}{\mathcal{L}}
\newcommand{\ie}{\textit{i.e.}}
\newcommand{\eg}{\textit{e.g.}}

\newcommand*{\mythead}[1]{\multicolumn{1}{c}{\bfseries #1}}

\newcommand{\reldec}{\textsc{RelDec}}
\newcommand{\relie}{\textsc{RelIE}}
\newcommand{\relu}{\text{ReLU}}
\newcommand{\enc}{\text{enc}}
\newcommand{\dec}{\text{dec}}
\newcommand{\ieig}{\hat{\text{IE}}_{\text{ig}}}
\newcommand{\ieigi}{\hat{\text{IE}}_{\text{ig},i}}
\newcommand{\compar}{$\leftrightarrow$}
\newcommand{\fv}{\mathbf{f}}
\newcommand{\bv}{\mathbf{b}}
\newcommand{\xv}{\mathbf{x}}
\newcommand{\xhatv}{\mathbf{\hat{x}}}
\newcommand{\wencc}{W^{\text{c}}_{\enc}}
\newcommand{\wdecc}{W^{\text{c}}_{\dec}}
\newcommand{\wdecci}{W^{\text{c}}_{\dec,i}}
\newcommand{\wdeccih}[1]{W^{#1}_{\dec,i}}

\newcommand{\bdecc}{\bv^{\text{c}}_{\dec}}
\newcommand{\benc}{\bv_{\enc}}

\makeatletter
  \xpretocmd\@declaredcolor
    {\my@hack@definetempcolor{#1}}
    {}{\PatchFailed}

  \xpretocmd\XC@col@rlet
    {\my@hack@definetempcolor{#4}}
    {}{\PatchFailed}

  \def\my@hack@definetempcolor#1{%
    \expanded{\my@hack@@definetempcolor{\zap@space#1 \@empty}}%
  }%

  \protected\def\my@hack@@definetempcolor#1{%
    \in@|{#1}%
    \ifin@
      \@ifundefinedcolor{#1}{%
        \my@hack@splitcolor#1\@nil{\xglobal\definecolor}%
      }{}%
    \fi
  }

  \def\my@hack@splitcolor#1|#2\@nil#3{%
    #3{#1|#2}{#1}{#2}%
  }
\makeatother

\definecolor{myred}{HTML}{E11D3F}
\definecolor{myorange}{HTML}{F04A00}
\definecolor{myblue}{HTML}{1974D2}
\definecolor{mypink}{HTML}{DA498D}
\definecolor{darkgreen}{HTML}{009B55}
\definecolor{mypink2}{HTML}{9823FF}

\title{Crosscoding Through Time: Tracking Emergence \& Consolidation \\ Of Linguistic Representations Throughout LLM Pretraining}

\author{
 \textbf{Deniz Bayazit\textsuperscript{1}},
 \textbf{Aaron Mueller\textsuperscript{2}},
 \textbf{Antoine Bosselut\textsuperscript{1}}
\\
 \textsuperscript{1}EPFL,
 \textsuperscript{2}Boston University
\\
 \small{
   \textbf{Correspondence:} \texttt{\{deniz.bayazit,antoine.bosselut\}@epfl.ch}
 }
}

\begin{document}

\maketitle
\begin{abstract}
Large language models (LLMs) learn non-trivial abstractions during pretraining, such as detecting irregular plural noun subjects. However, because traditional evaluation methods (\eg{}, benchmarking) fail to reveal how models acquire these concepts and capabilities, it is not well understood when and how these specific linguistic abilities emerge. To bridge this gap and better understand model training at the concept level, we use sparse crosscoders to discover and align features across model checkpoints. Using this approach, we track the evolution of linguistic features during pretraining. We train crosscoders between open-sourced checkpoint triplets with significant performance and representation shifts, and introduce a novel metric, Relative Indirect Effects (\relie{}), to trace training stages at which individual features become causally important for task performance. We show that crosscoders can detect feature emergence, maintenance, and discontinuation during pretraining. Our approach is architecture-agnostic and scalable, offering a promising path toward more interpretable and fine-grained analysis of representation learning throughout pretraining.\footnote{The code, crosscoders, and annotations are available at \\
{\url{https://github.com/bayazitdeniz/crosscoding-through-time}}}
\end{abstract}
\section{Introduction}

\begin{figure}[t]
\centering
\includegraphics[width=0.95\linewidth]{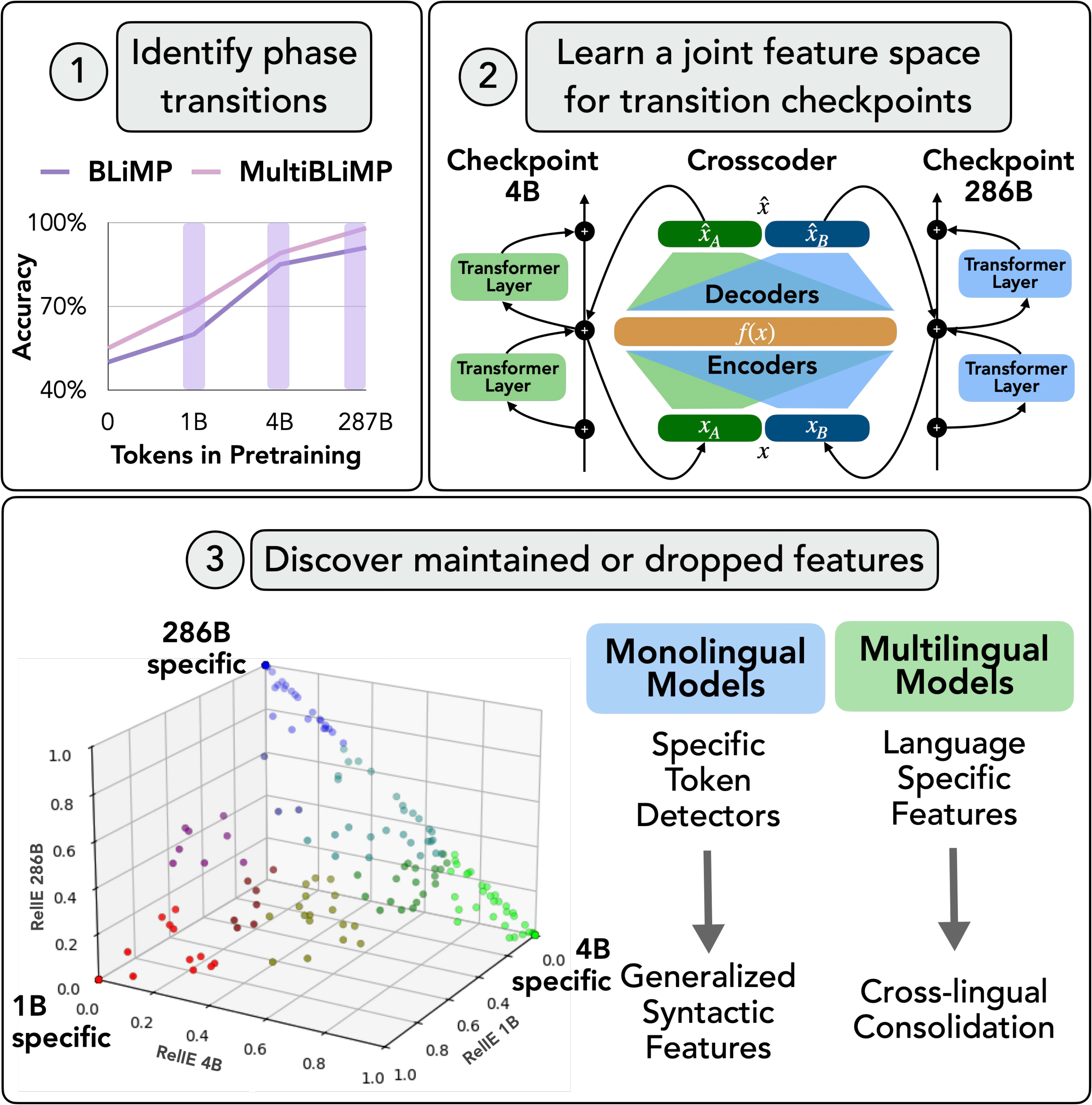}
\caption{\textbf{Capturing the evolution of features.} Given a task, our pipeline selects the relevant checkpoints during pretraining, learns a joint feature space with crosscoders, and then analyzes causal feature importance across checkpoints. This allows to understand how models learn, maintain, or unlearn particular representations over time.}
\label{fig:first-fig}
\vspace{-1.25em}
\end{figure}

Among the foundational advances in deep learning is the ability to learn useful internal features through gradient-based optimization, rather than relying on hand-crafted representations \citep{rumelhart1986learning}. This principle underlies much of the success of modern large language models (LLMs), where learned features can capture complex linguistic patterns during training \citep{manning2020emergent}. However, this unstructured learning comes at the expense of interpretability \citep{mueller2024questrightmediatorhistory}, motivating new methods that measure whether particular concept representations are learned by LMs. 

In particular, we lack a clear understanding of when and how specific linguistic abilities emerge during pretraining---a gap that, if bridged, would allow us to better understand LLM pretraining at the concept level. Common methods to estimate the acquisition of a concept include measuring the performance on tasks that act as proxies for the concept \citep{olsson2022context, chen2024sudden}, and identifying changes in the model's activation or parameter spaces \citep{wu-etal-2020-similarity}. However, these measures only reveal \emph{when} changes occur and fail to shed light on the mechanism by which a model internalizes particular linguistic concepts (\eg{}, subject--verb agreement; \citealp{lovering2021predicting, bunzeck-zarriess-2024-fifty, kangaslahti2025hiddenbreakthroughs}).

Recently, sparse autoencoders (SAEs) have been adopted to study \emph{how} such linguistic concepts are represented by models. SAEs project a model’s dense internal representation at particular layers onto large, sparsely activated feature spaces \citep{bricken2023monosemanticity,huben2024sparse}, thereby discretizing activations into linear combinations of one‑dimensional features. However, using SAEs to better understand the evolution of concepts during pretraining would require training unique SAEs for all checkpoints. These separately learned sparse feature spaces would preclude direct feature comparisons across training stages. To address these limitations, sparse crosscoders were introduced to learn a single \emph{joint} feature space across layers or models simultaneously \citep{lindsey2024crosscoder}. This framework provides a structured lens for analyzing how linguistic concepts evolve over checkpoints: shared features indicate concepts that persist, while unique features reveal concepts that emerge or disappear. However, prior work has only used crosscoders to study features that arise during post-training \citep{minder2025robustlyidentifyingconcepts,baek2025distilled}.

In this work, we use crosscoders to track the evolution of syntactic concept representations across pretraining checkpoints, using the pipeline illustrated in Fig.~\ref{fig:first-fig}. First, we learn crosscoders across checkpoint triplets showing behavioral and representational shifts on benchmarks such as BLiMP \cite{warstadt-etal-2020-blimp-benchmark}. We compare checkpoints from the same training run so that representational changes can be attributed to pretraining dynamics rather than to differences in tokenizer, data mixture, or objective. Then, we introduce the Relative Indirect Effect (\relie{}) metric to causally quantify per-feature attribution over training checkpoints and annotate the role of the features. We validate \relie{} through ablation and interpretability studies, assessing whether it accurately traces how and when features gain or lose task relevance.

We show that pairing crosscoders with our \relie{} metric enables us to pinpoint linguistic concept representations at individual checkpoints and trace their development over time. This architecture-agnostic framework---validated on Pythia, BLOOM, and OLMo---scales easily to billion-parameter models. Qualitatively, we find that LLMs progressively build higher-level abstractions, as evidenced in how token and language-specific concepts gradually become abstracted into more universal concepts.
\section{Related Work}

\paragraph{Language Model Interpretability} 
There has recently been significant progress in scaling unsupervised interpretability methods, including dictionary learning \citep{bricken2023monosemanticity,huben2024sparse}, circuit discovery \citep{wang2023interpretability,conmy2023circuit, bayazit-etal-2024-discovering}, and work that combines the two \citep{marks2025featcircuits}. These have been instrumental in revealing \textit{how} a final checkpoint performs tasks like subject--verb agreement \citep{marks2025featcircuits}, parenthesis matching \citep{huben2024sparse}, garden-path sentence processing \citep{hanna-mueller-2025-incremental}, and crosslingual morphosyntactic generalization \citep{brinkmann-etal-2025-large}. However, they offer limited insight into \emph{when} specific concepts emerge. Crosscoders \citep{lindsey2024crosscoder} address this gap by mapping joint feature spaces between models (\eg{}, pretrained vs. instruction-tuned; \citealp{minder2025robustlyidentifyingconcepts,baek2025distilled}). We extend this approach to pretraining checkpoints to trace concept-level feature evolution and uncover both \textit{when} and \textit{how} representations emerge.

\paragraph{Training Dynamics} A parallel line of work examines the learning trajectories of models via model performance, parameter shifts, and activation patterns across training steps \citep{saphra-lopez-2019-understanding,wu-etal-2020-similarity,kaplan2020scalinglaws,liu-etal-2021-probing-across}.
Some research aligns these dynamics with cognitive signals (\eg{}, brain activity; \citealp{nakagi2025triplephase, alkhamissi-etal-2025-llmlanguagenetwork, alkhamissi2025languagecognition, constantinescu-etal-2025-investigating}), or tracks knowledge acquisition over time \citep{liu-etal-2021-probing-across, ou2025llmsacquirenewknowledge, cao-etal-2024-retentive, zucchet2025languagemodelslearnfacts}. Such studies shed light on what general internal changes occur, but they do not enable precise concept-level claims and rarely tie these changes back to discrete or human‐interpretable conceptual representations.

More recently, \citet{kangaslahti2025hiddenbreakthroughs} introduce POLCA, a method that analyzes training loss patterns to uncover hidden phase transitions among conceptually similar data samples. While POLCA reveals when a concept emerges through \emph{loss dynamics}, our approach aims to trace how such a concept's role evolves over time by using the model's \emph{hidden states}. Together, these complementary paradigms deepen our understanding of how pretraining shapes model behavior. 

\section{Preliminaries}
\label{sec:preliminaries}

\paragraph{Crosscoders}
SAEs\footnote{We provide a formal definition of SAEs in Appendix~\ref{sec:appendix_premilinaries}.} learn mappings from activations of model layers to feature spaces $\fv$. Consequently, these mappings are unique to the corresponding model layer activations that are used as input during training, and cannot be used to disentangle what concepts might be shared or unique across different activation spaces (such as those in different model layers, or from different model checkpoints). Sparse crosscoders instead learn a joint feature space for activations from multiple sources, \eg{}, from multiple checkpoints, denoted $C=\{\,c_1,c_2,\dots\}$. Crosscoders introduce three key modifications to the SAE paradigm: (1) dedicated encoder and decoder weights per source $c \in C$ ($\wdecc$, $\bdecc$ and $\wencc$) to capture source-specific concepts;\footnote{Note that $\benc$ is shared across checkpoints.} (2) a joint reconstruction loss that is averaged across the different sources $\xv_{c}$ and their reconstructions $\xhatv_{c}$; (3) an aggregated sparsity penalty summed across the sources to encourage the inclusion of both shared and unique features in the joint feature space. The final loss is then:
\begin{align} 
\fv &=\relu \Big( \sum\nolimits_{c \in C} \wencc \xv_{c} + \benc\Big)  \\ 
\xhatv_{c} &= \wdecc \fv + \bdecc \\
\begin{split}
\Loss &= \sum\nolimits_{c \in C} \lVert \xv_{c} - \xhatv_{c}\rVert_{2}^{2} \\
& \quad + \sum\nolimits_{c \in C} \sum\nolimits_{i} \fv_i \lVert \wdecci \rVert_{2}, 
\end{split}
\end{align}
\noindent where $i$ indexes a particular feature in $\fv$ and the column $\wdecc$ that scales $\fv_i$ when reconstructing $\xhatv_c$.

\paragraph{Measuring feature changes}
To determine whether a crosscoder feature $\fv_i$ is unique to a particular checkpoint or shared between checkpoints, prior work proposes the relative decoder norm (\reldec) $\in [0,1]$ \cite{lindsey2024crosscoder}, computed per feature $\fv_i$:
\begin{equation} \label{eq:reldec}
\reldec_{i} = \frac{{\left\lVert \wdeccih{\text{c}_2}\right\rVert_2}}{\sum\limits_{c \in \{c_1, c_2\}}{\left\lVert \wdeccih{\text{c}} \right\rVert_2}}
\end{equation}
\noindent For each checkpoint $c$ and dictionary feature $\fv_i$, the $\ell_2$ norm is computed across the hidden model activation dimension that we aim to reconstruct. Then, the norms are scaled across features by their strength per model to find which feature is more specific to a given model, or shared. Values closer to 0 mean the feature is more present in $c_1$; those closer to 1 mean the feature is more present in $c_2$.

\paragraph{Indirect Effect}
In the tasks we study, for a given prefix, \ie{} a token set $x$, a single token identifies a correct ($t_{\mathrm{correct}}$) or wrong ($t_{\mathrm{wrong}}$) completion. To quantify each hidden unit’s contribution to the correct completion---whether it is a neuron (\ie{}, a dimension in the residual/layer output vector) or a crosscoder feature---we compute its indirect effect (IE) at each checkpoint by zero-ablating  the feature (\ie{}, setting its activation to 0) and measuring the change in a selected metric. To measure the significance of the feature towards the correct model behavior we compute the following log-probability difference as the primary metric $m$:
$$
 m(x) = \log p(t_{\mathrm{wrong}} \mid x) - \log p(t_{\mathrm{correct}} \mid x)
$$
Specifically, IE is defined as the difference in $m(x)$ after and before the ablation $\mathbf{a}_{\text{patch}} $ \citep{pearl2001indirect}:
\begin{equation}
\label{eq:ieig}
\text{IE}(m; \mathbf{a}; x) = m\bigl(x \mid \operatorname{do}(\mathbf{a} = \mathbf{a}_{\text{patch}})\bigr) - m(x),
\end{equation}
where the $\operatorname{do}$ operator replaces the original activation $\mathbf{a}$ with $\mathbf{a}_{\text{patch}}$. Intuitively, a positive IE indicates that ablating the unit pushes the model’s prediction away from the correct class, while a negative IE means the ablation reinforces the correct prediction. In our work, we use integrated gradients ($\ieig$; \citealp{sundararajan2017axiomaticattr, marks2025featcircuits}) to approximate the IE of crosscoder features and use zero-ablation as patching. For more details on our IE implementation, see Appendix~\ref{sec:appendix_ie}. 
\section{Methodology}
To understand how model representations evolve across training, we aim to identify which features emerge, persist, or disappear over time. This requires attributing representations to \emph{specific} training checkpoints, a task we refer to as checkpoint representation attribution. We tackle this in three steps: 
(1) Identify the critical checkpoints via performance and activation correlation analyses;
(2) Learn a crosscoder between these critical checkpoints;
(3) Attribute features using the Relative Indirect Effect (see Eq.~\ref{eq:relie2way} below), and track emerging, maintained, or vanishing representations.

\begin{figure*}[th]
    \centering
    \begin{subfigure}{0.3\linewidth}
      \centering
      \includegraphics[width=\textwidth]{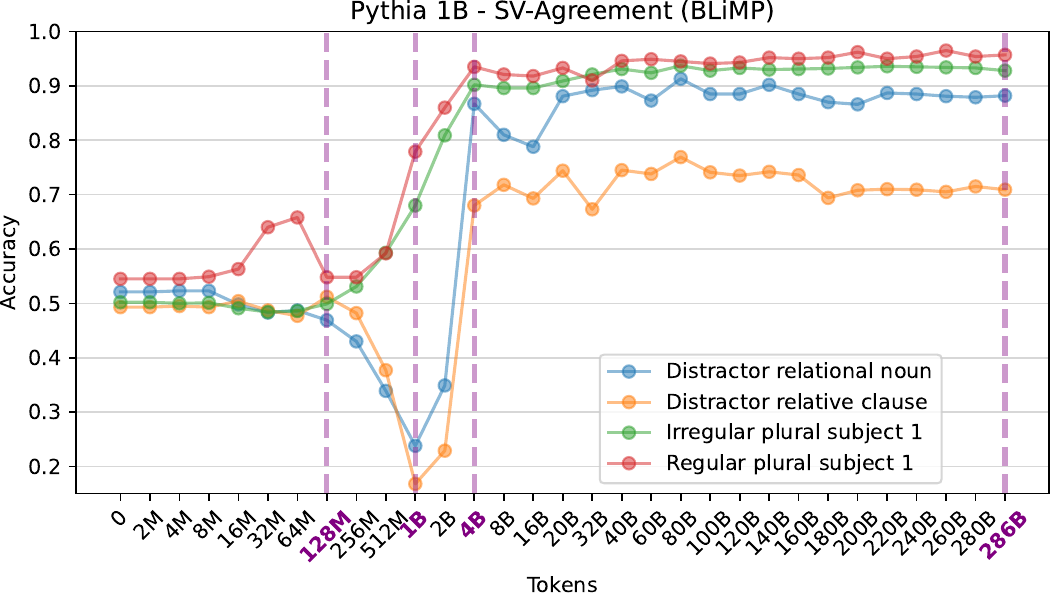}
    \end{subfigure}
    \begin{subfigure}{0.3\linewidth}
      \centering
      \includegraphics[width=\textwidth]{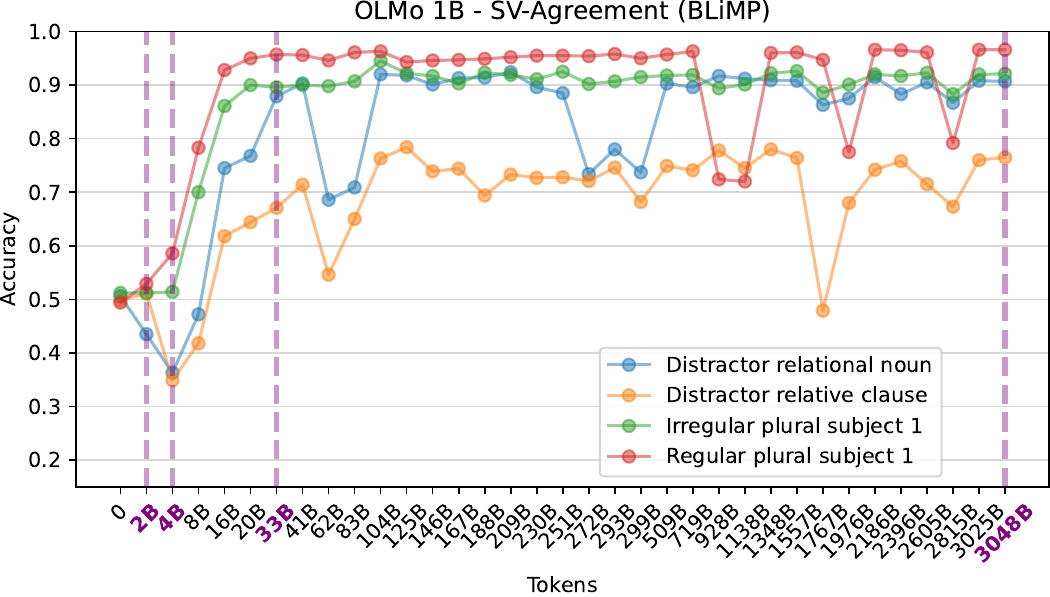}
    \end{subfigure}
    \begin{subfigure}{0.3\linewidth}
      \centering
      \includegraphics[width=\textwidth]{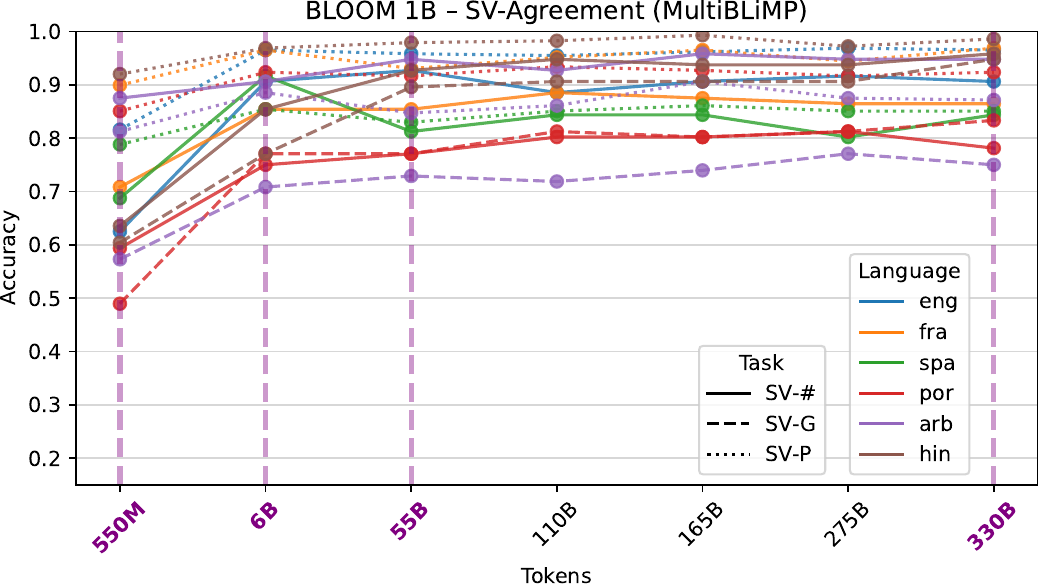}
    \end{subfigure}

    \vspace{3pt} 
    \begin{subfigure}{0.3\linewidth}
      \centering
      \includegraphics[width=\textwidth]{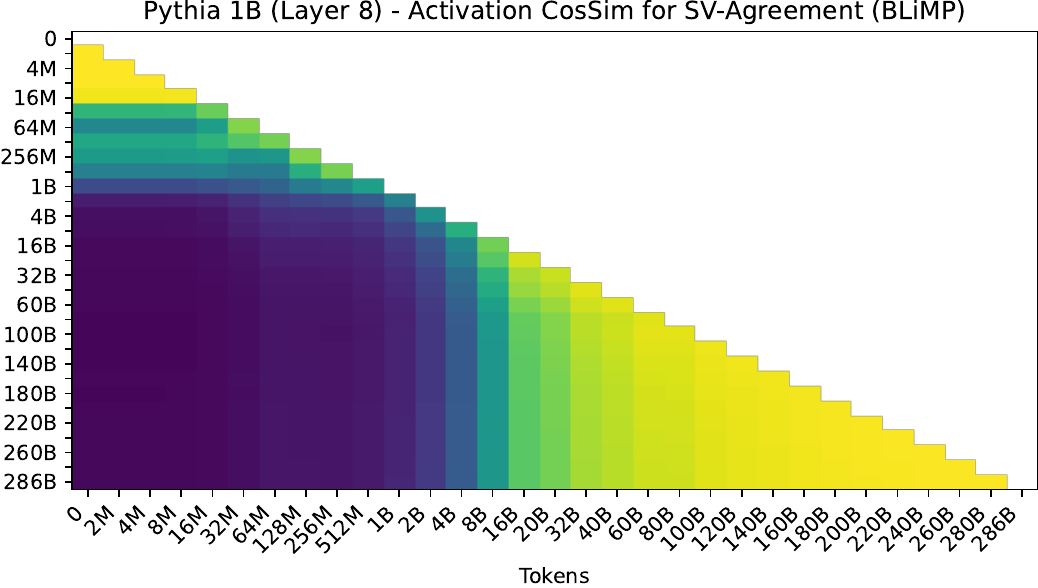}
    \end{subfigure}
    \begin{subfigure}{0.3\linewidth}
      \centering
      \includegraphics[width=\textwidth]{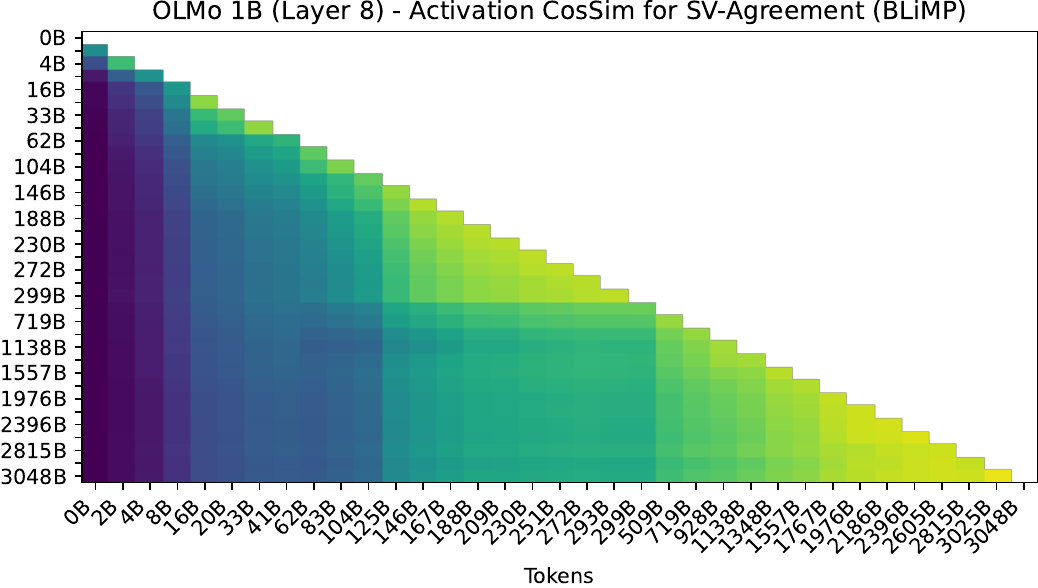}
    \end{subfigure}
    \begin{subfigure}{0.3\linewidth}
      \centering
      \includegraphics[width=\textwidth]{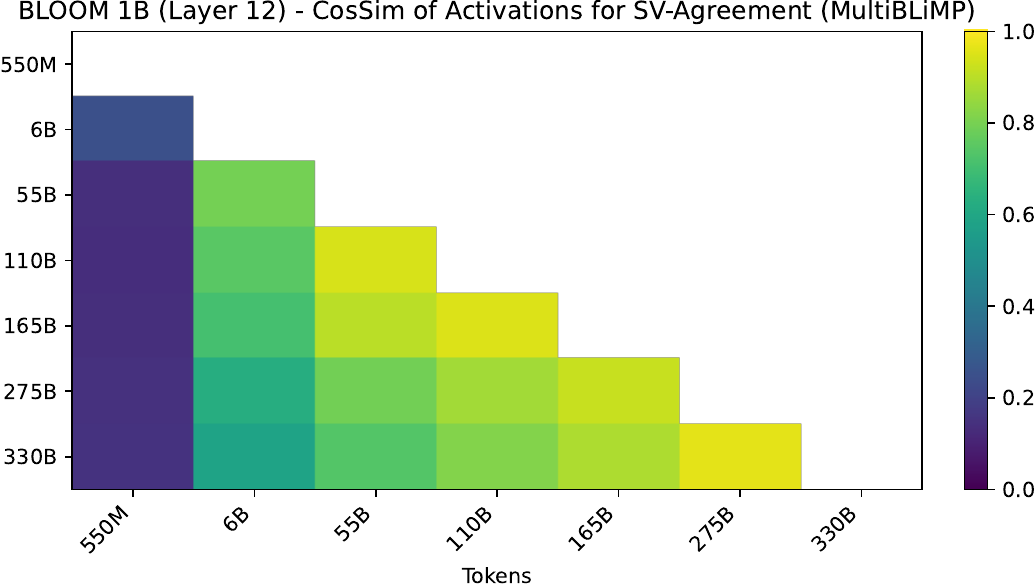}
    \end{subfigure}
    \caption{\textbf{Checkpoint selection with task performance (top) and average middle-layer activation cosine similarity (bottom).} The Pythia-1B (left) and OLMo-1B (middle) performance and activations patterns are calculated using BLiMP whereas BLOOM-1B (right) uses MultiBLiMP. All columns have the same $x$-axis (number of training tokens). We highlight checkpoints identified as critical in purple vertical lines. While some activation shifts align with performance jumps, others reveal continued representational change even after accuracy plateaus.}
    \label{fig:checkpoint-analysis}
\end{figure*}

\paragraph{Phase Transition Identification}
Building on prior work that flags sudden jumps (\ie, phase transitions) in validation accuracy, loss, or activation-pattern similarity \citep{wu-etal-2020-similarity, chen2024sudden, nakagi2025triplephase}, we track two signals from different checkpoints in tandem: (1) each checkpoint’s accuracy on the target task, and (2) the pairwise correlation of mid-layer activations across all checkpoints, averaged across task inputs. 
We focus on mid-layer activations because prior work has shown that they capture higher-level linguistic and cross-lingual abstractions \cite{tenney-etal-2019-bert, liu-niehues-2025-middle, brinkmann-etal-2025-large}, while earlier and later layers are more closely tied to input processing and output prediction \cite{lad2024robustnessofllms, csordas2025languagemodelsusedepth}. This makes the mid-layer activations a natural site for analyzing how such representations evolve over training.
By plotting accuracy and the mid-layer activation similarity across training steps (later shown in Fig.~\ref{fig:checkpoint-analysis}), we identify when the model undergoes representational shifts.

\paragraph{Crosscoder Training for Checkpoint Evolution} 
We train crosscoders under two regimes to trace the evolution of linguistic representations. First, we conduct triplet comparisons of phase-transition checkpoints (denoted by the number of tokens they have been trained on, \eg{}, 1B \compar{} 4B \compar{} 286B) to gauge which features persist between phases.\footnote{A notable challenge is that very early checkpoints often resist sparse mapping and accurate reconstruction; prior work shows SAEs falter on fully random models \citep{karvonen2024boardgame}, but their behavior on \textit{partially-trained} checkpoints remains underexplored. We verify that our crosscoders remain robust even when incorporating these early checkpoints (\S\ref{subsec:learnability}).} Then, we analyze pairwise comparisons to verify that the takeaways from triplet comparisons are consistent with pairwise observations. In each setting, we also include the final training checkpoint of a particular model to see which features are present in the model’s fully trained state.

\paragraph{Feature Selection \& Annotation with \relie{}}
The task-agnostic \reldec{} (Eq.~\ref{eq:reldec}) separates checkpoint-specific features from shared ones by comparing \emph{every} feature in the dictionary. While this yields a broad, task-agnostic view, it makes it difficult to isolate the features that actually drive performance on a target task and to interpret their role. Instead, we compute each feature’s Indirect Effect (IE) per checkpoint at each token step using integrated gradients (Eq.~\ref{eq:ieig}), which directly quantifies a feature’s contribution for the correct behavior. We then define the Relative Indirect Effect (\relie{}) as the ratio of the absolute approximated IEs ($\ieig$) for checkpoints $c_1$ and $c_2$ for each feature $\fv_i$, following the same normalization as Eq.~\ref{eq:reldec} but applied to IEs:\footnote{Ablations in Appendix~\ref{sec:appendix_ablation} show that by focusing on task‐relevant signal rather than the entire feature set, \relie{} uncovers more meaningful task-specific features.}
\begin{equation} \label{eq:relie2way}
\relie_{\text{2-way},i} = \frac{|\ieigi^{\text{c}_2}|}{|\ieigi^{\text{c}_1}| + |\ieigi^{\text{c}_2}|}
\end{equation}
For three-checkpoint crosscoders, we compute a one-versus-all \relie{} via:
\begin{equation} \label{eq:relie3way}
\relie_{\text{3-way},i} = \frac{\bigl(|\ieigi^{\text{c}_1}|,|\ieigi^{\text{c}_2}|,|\ieigi^{\text{c}_3}|\bigr)}
{\sum\nolimits_{c \in \{c_1, c_2, c_3\}} \bigl|\ieigi^{\text{c}}\bigr|}
\end{equation}
We then select each checkpoint’s top-10 IE features for the particular task, annotate them by the pretraining sequences that maximally activate them, and use \relie{} to trace how their task relevance shifts across checkpoints. 
\section{Experimental Setup}
\label{sec:setup}

\paragraph{Models \& Crosscoders}  
We evaluate three open-source LLM families with publicly released checkpoints: Pythia 1B \cite{biderman2023pythia}, OLMo 1B \cite{groeneveld-etal-2024-olmo}, and BLOOM 1B \cite{bigscience2023bloom}. Pythia’s dense checkpoint logging lets us understand early linguistic feature emergence more precisely; OLMo’s extended training helps us study feature maintenance over longer pretraining; and BLOOM’s multilingual corpus allows us to trace crosslingual representation development. Following prior SAE work at similar model scale \cite{lieberum-etal-2024-gemma}, our crosscoders use a dictionary size of $2^{14}$ features. We train crosscoders on the output hidden states extracted from the middle transformer block of each checkpoint. Additional details are provided in Appendix~\ref{sec:appendix_model}~\&~\ref{sec:appendix_crosscoders}.

\paragraph{Crosscoder Training Datasets}
Guided by the principle that a system’s behavior reflects its training distribution \citep{mccoy-etal-2024-embers}, we train each crosscoder on a subset of its model’s original pretraining data subsampled to 400M tokens. For Pythia, we sample from the Pile \cite{gao2020pile}; for OLMo, we sample from Dolma \citep{soldaini-etal-2024-dolma}; and for BLOOM, we subsample mC4 \citep{xue-etal-2021-mt5} in proportion to the top ten languages represented in ROOTS \citep{laurencon2023roots}.

\paragraph{Linguistic Tasks} 
To chart the acquisition of subject--verb agreement representations in LLMs, we use the BLiMP \citep{warstadt-etal-2020-blimp-benchmark}, MultiBLiMP \citep{jumelet2025multiblimp}, and CLAMS \citep{mueller-etal-2020-cross} benchmarks. 
These benchmarks are especially useful for our causal analysis because they are built around controlled minimal pairs, allowing us to compare model behavior on sentence pairs that differ only in the targeted grammatical phenomenon. At the same time, they cover a range of agreement-related subtasks that vary in grammatical case, language, and difficulty (\eg{}, BLiMP includes verb agreement cases with regular vs.\ irregular plural noun subjects).
We preprocess them by finding examples where the difference between a correct and wrong completion is determined by a single token for a given prefix. See Appendix~\ref{sec:appendix_dataset} for processing details and examples. 
\section{Crosscoder Learnability for Intermediate Checkpoints}

\subsection{Checkpoint Selection via Phase Transition Identification}

Before training crosscoders, we identify checkpoints where phase transitions in model behavior or representations occur. To uncover phase transitions, we track subject--verb agreement accuracy (BLiMP, MultiBLiMP) and the cosine similarity of middle‐layer activations across checkpoints for Pythia-1B, OLMo-1B, and BLOOM-1B (Fig.~\ref{fig:checkpoint-analysis}).\footnote{All observed average pairwise cosine similarity values were non-negative, so we restrict the heatmap range to $[0,1]$.}

In Pythia-1B, a significant phase transition unfolds from 128M to 4B tokens: accuracy on BLiMP vaults from near‐chance (\(\sim\)50\%) to above 90\%, and the activation similarity heatmap shows a significant change compared to earlier near-random checkpoints. These indicators signal the emergence of syntactic representations. A smaller inflection at 1B hints that certain subtasks (\eg{}, agreement with irregular plural subjects) are learned before more complex ones at 4B (\eg{}, agreement with distractor clauses), after which performance remains stable until 286B. Hence, we study checkpoints that have been trained on \{128M, 1B, 4B, 286B\} tokens.

OLMo-1B follows a more staggered trajectory: an initial milestone at 2B boosts accuracy and activation similarity, followed by a second adjustment phase at 4B, and a consolidation near 33B tokens. After 33B, while accuracy mostly plateaus, activations continue refining through 3T, indicating ongoing representational changes. Thus we select the \{2B, 4B, 33B, 3T\} checkpoints for OLMo-1B.

Applying the same analysis to BLOOM-1B on MultiBLiMP, while accuracy jumps for all subtasks at 6B, the extent differs for particular languages like English vs.\ Arabic. Beyond 55B tokens, performance plateaus but continues to undergo subtle refinement of the activation space until 341B, hence the choice of \{550M, 6B, 55B, 341B\} checkpoints.

\begin{table*}[!ht]
\centering
\resizebox{0.9\linewidth}{!}{%
\begin{tabular}{crll}
\toprule
\mythead{RelIE} & \mythead{FeatID} & \mythead{Interpreted Function} & \mythead{Top Activating Sequence} \\
\midrule
\multicolumn{4}{l}{\bfseries 1B-4B shared} \\
\midrule
$[0.53,\;0.33,\;0.15]$ & 1067 & \makecell[l]{Detects subtoken \emph{-ans} in various contexts} & 
    \makecell[l]{%
      \sethlcolor{white}\textcolor{black}{\hl{... pick‑ups, v}}%
      \sethlcolor{RGB|22,22,255}\textcolor{white}{\hl{ans}}%
      \sethlcolor{white}\textcolor{black}{\hl{, and larger vehicles such ...}}%
    } \\
$[0.45,\;0.41,\;0.14]$ & 4897 & \makecell[l]{Detects subtoken \emph{-ists},\\[-0.7ex](\eg{}, protagonist, capitalist, pharmacist)} & 
    \makecell[l]{%
      \sethlcolor{white}\textcolor{black}{\hl{... a conspiracy by elit}}%
      \sethlcolor{RGB|22,22,255}\textcolor{white}{\hl{ists}}%
      \sethlcolor{white}\textcolor{black}{\hl{ within government and big business ...}}%
    } \\
\multicolumn{4}{l}{\bfseries 1B-286B shared} \\
\midrule
$[0.52,\;0.01,\;0.46]$ & 3852 & \makecell[l]{Detects \emph{man} subtoken,often the singular noun\\[-0.7ex]but sometimes not (\eg{}, manned aircraft)} & 
    \makecell[l]{%
      \sethlcolor{white}\textcolor{black}{\hl{... bad omens: The }}%
      \sethlcolor{RGB|22,22,255}\textcolor{white}{\hl{man}}%
      \sethlcolor{white}\textcolor{black}{\hl{ in charge of the reactor ...}}%
    } \\
$[0.55,\;0.10,\;0.34]$ & 7489 & Detects singular \emph{woman} noun & 
    \makecell[l]{
      \sethlcolor{white}\textcolor{black}{\hl{... other people –; including a }}%
      \sethlcolor{RGB|69,69,255}\textcolor{white}{\hl{woman}}%
      \sethlcolor{white}\textcolor{black}{\hl{ with far too many cats ...}}%
    } \\
\multicolumn{4}{l}{\bfseries 4B specific} \\
\midrule
$[0.00,\;1.00,\;0.00]$ & 11274 & \makecell[l]{Multi-word noun or compound noun detector\\[-0.7ex]in particular for scientific writing} & 
    \makecell[l]{%
      \sethlcolor{white}\textcolor{black}{\hl{...}}%
      \sethlcolor{RGB|180,180,255}\textcolor{white}{\hl{ two}}%
      \sethlcolor{RGB|180,180,255}\textcolor{white}{\hl{-}}%
      \sethlcolor{RGB|150,150,255}\textcolor{white}{\hl{dimensional}}%
      \sethlcolor{white}\textcolor{black}{\hl{ }}%
      \sethlcolor{RGB|100,100,255}\textcolor{white}{\hl{pair}}%
      \sethlcolor{RGB|100,100,255}\textcolor{white}{\hl{ correlation }}%
      \sethlcolor{RGB|22,22,255}\textcolor{white}{\hl{function}}%
      \sethlcolor{white}\textcolor{black}{\hl{ and the }}%
      \sethlcolor{RGB|180,180,255}\textcolor{white}{\hl{structure}}%
      \sethlcolor{white}\textcolor{black}{\hl{ }}%
      \sethlcolor{RGB|22,22,255}\textcolor{white}{\hl{factor}}%
    } \\
$[0.02,\;0.68,\;0.30]$ & 10523 & \makecell[l]{Detects plural nouns depicting humans\\[-0.7ex](\eg{}, people, students, bloggers)} & 
    \makecell[l]{%
      \sethlcolor{white}\textcolor{black}{\hl{Most }}%
      \sethlcolor{RGB|150,150,255}\textcolor{white}{\hl{older}}%
      \sethlcolor{white}\textcolor{black}{\hl{ }}%
      \sethlcolor{RGB|22,22,255}\textcolor{white}{\hl{scientists}}%
      \sethlcolor{white}\textcolor{black}{\hl{ said such a thing is ...}}%
    } \\
\multicolumn{4}{l}{\bfseries 286B specific} \\
\midrule
$[0.08,\;0.15,\;0.77]$ & 14228 & \makecell[l]{Multi-word named entity and title detectors\\[-0.7ex](\eg{}, proper nouns, locations etc.)} &  
    \makecell[l]{%
      \sethlcolor{RGB|180,180,255}\textcolor{white}{\hl{Buick}}%
      \sethlcolor{white}\textcolor{black}{\hl{ }}%
      \sethlcolor{RGB|150,150,255}\textcolor{white}{\hl{Lesabre}}%
      \sethlcolor{white}\textcolor{black}{\hl{ }}%
      \sethlcolor{RGB|100,100,255}\textcolor{white}{\hl{Ownership}}%
      \sethlcolor{white}\textcolor{black}{\hl{ }}%
      \sethlcolor{RGB|22,22,255}\textcolor{white}{\hl{Costs}}%
      \sethlcolor{white}\textcolor{black}{\hl{ There}}%
    } \\
$[0.00,\;0.18,\;0.82]$ & 6746 & \makecell[l]{Detects deverbal nouns and\\[-0.7ex] nominalizations formed from verbs} & 
    \makecell[l]{%
      \sethlcolor{white}\textcolor{black}{\hl{Though my }}%
      \sethlcolor{RGB|22,22,255}\textcolor{white}{\hl{reactions}}%
      \sethlcolor{white}\textcolor{black}{\hl{ tend to be tepid ...}}%
\\[-0.7ex]
  \sethlcolor{white}\textcolor{black}{\hl{However its }}%
  \sethlcolor{RGB|22,22,255}\textcolor{white}{\hl{performance}}%
  \sethlcolor{white}\textcolor{black}{\hl{ degrades with ...}}%
}
\\
$[0.00,\;0.01,\;0.99]$ & 14623 & \makecell[l]{Detects prepositions} & 
    \makecell[l]{%
      \sethlcolor{white}\textcolor{black}{\hl{...so that an inclusion bias }}%
      \sethlcolor{RGB|22,22,255}\textcolor{white}{\hl{due}}%
      \sethlcolor{RGB|150,150,255}\textcolor{white}{\hl{ to}}%
      \sethlcolor{white}\textcolor{black}{\hl{ restricted use }}%
      \sethlcolor{RGB|22,22,255}\textcolor{white}{\hl{in}}%
      \sethlcolor{white}\textcolor{black}{\hl{ specific}}%
    } \\
\bottomrule
\end{tabular}
}%
\caption{\textbf{Subset of 3‑way Crosscoder Annotations for Pythia‑1B | 1B\compar{}4B\compar{}286B tokens.}  
\relie{} gives the one‑versus‑all attribution vector; Interpreted function describes any detected linguistic role. Rows are grouped by whether features are unique to one checkpoint, shared between two, or common to all. Top activating sequence shows an example yielding a high activation, where color density shows the activation intensity for individual tokens. Features evolve from token-specific detectors to group-level abstract concepts (\eg{}, deverbal noun detectors), even after accuracy plateaus.}
\label{tab:3way_annotation_pythia_subset}
\vspace{-0.5em}
\end{table*}

 \begin{figure}[t]
    \centering
    \includegraphics[width=1.0\linewidth]{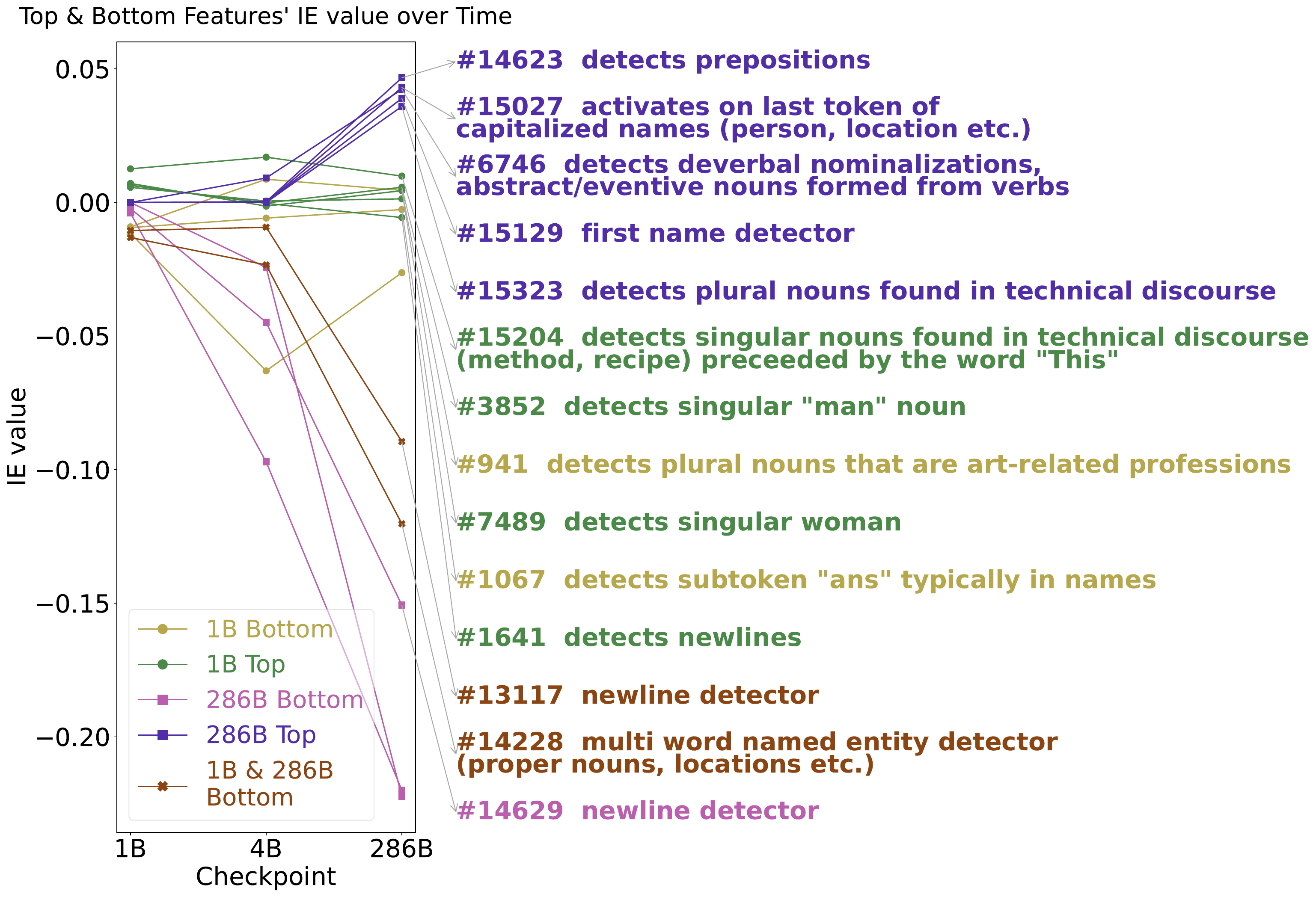}
    \caption{\textbf{IE evolution of Top-5 \& Bottom-5 IE Features for Pythia checkpoints 1B \& 286B.} IEs are calculated using BLiMP subject--verb agreement tasks. Missing annotation means the feature was not interpretable. We observe that low-level or uninterpretable features fade over time, while high-level grammar detectors emerge and strengthen by 286B.}
    \label{fig:ie-evolution-pythia}
    \vspace{-1em}
\end{figure}

\subsection{Crosscoder Learnability}
\label{subsec:learnability}
We train crosscoders on pairs and triplets of checkpoints using subsampled model training data (\S\ref{sec:setup}). As seen in Table~\ref{tab:crosscoder-stats} (Appendix~\ref{sec:appendix_crosscoders}), across three random seeds and checkpoints at different training stages, crosscoders consistently reconstruct intermediate activations with little increase in cross entropy loss compared to the original loss (mostly $\Delta$CE < 0.2), even for earlier checkpoints that are trained on an order-of-magnitude less data. Larger differences in tokens used for pretraining lead to more dead features (\ie{}, hidden units in the joint feature space that never get activated) and slightly higher $\Delta$CE (\(\sim\)0.35), but they remain largely recoverable.

\section{Emergence of Agreement Generalizations in Monolingual Models}

Having established crosscoder mappings at critical checkpoints, we examine whether monolingual models develop broader syntactic intuitions. Specifically, we investigate whether LLMs initially rely primarily on surface-level token matching mechanisms and then progressively internalize deeper grammatical abstractions. To evaluate this, we track how features evolve across training stages, annotating and quantifying the top and bottom IE features at each checkpoint.

\subsection{From Specific Token Detectors to High-Level Syntactic Features}

 \begin{figure*}[th]
    \centering
    \includegraphics[width=0.9\textwidth]{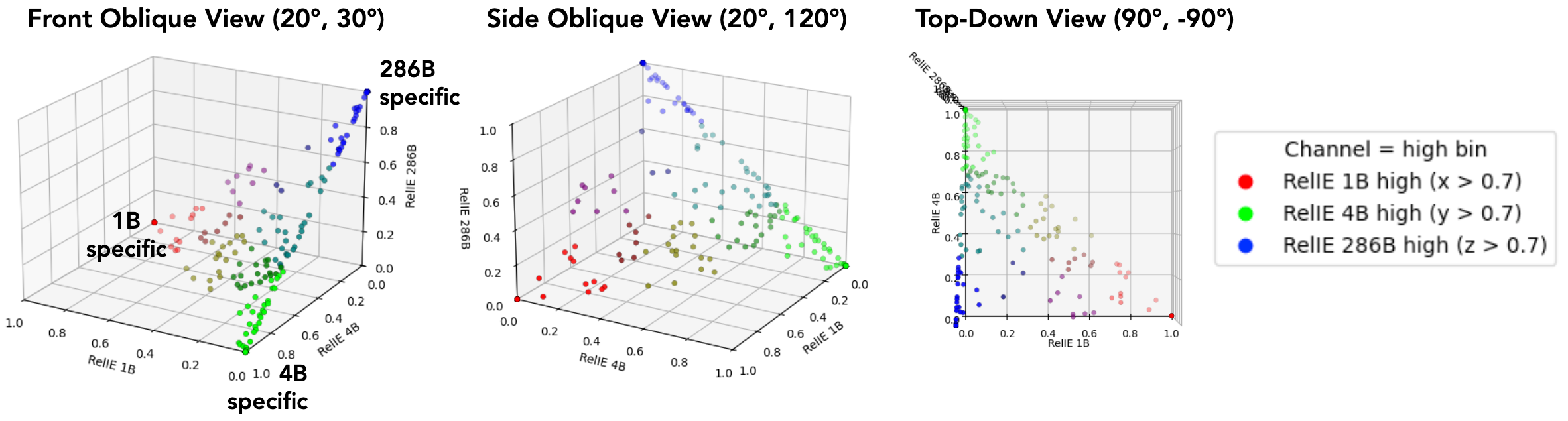}
    \caption{\textbf{\relie{} of Top-100 IE features on BLiMP for Pythia-1B checkpoints 1B, 4B, 286B.} Distinct clusters at each corner show attribution of checkpoint-specific features. 4B–286B share more abstract features, while 1B–286B overlaps are sparse and lower-level features.}
    \label{fig:monolingual-overlap}
\end{figure*}
\begin{table*}[!ht]
\centering
\resizebox{0.95\linewidth}{!}{%
\begin{tabular}{crcll}
\toprule
\mythead{RelIE} & \mythead{FeatID} & \mythead{Activated Languages} & \mythead{Interpreted Function} & \mythead{Top Activating Sequence}\\
\midrule

\multicolumn{4}{l}{\bfseries 6B specific} \\
\midrule
$[1.00,\;0.00,\;0.00]$ & 3672 & arb,eng,fra,hin,por,spa & \makecell[l]{Detects ellipsis and question/exclamation marks}  &  
    \makecell[l]{%
      \sethlcolor{white}\textcolor{black}{\hl{... ou mesurage sur place}}%
      \sethlcolor{RGB|22,22,255}\textcolor{white}{\hl{?}}%
      \sethlcolor{white}\textcolor{black}{\hl{Ville/Région :}}%
    } \\
$[0.78,\;0.14,\;0.07]$ & 7122 & eng & \makecell[l]{Main-verb head detector} &  
    \makecell[l]{%
      \sethlcolor{white}\textcolor{black}{\hl{... one day, Ananya }}%
      \sethlcolor{RGB|22,22,255}\textcolor{white}{\hl{opens}}%
      \sethlcolor{white}\textcolor{black}{\hl{ up about herself and her}}%
    } \\
$[0.62,\;0.17,\;0.20]$ & 15288 & fra,por,spa & \makecell[l]{Detects \emph{Ev}, \emph{ev}, \emph{év} subtokens\\[-0.7ex](only in latin languages)} & 
    \makecell[l]{%
      \sethlcolor{white}\textcolor{black}{\hl{... les résultats ont été moins }}%
      \sethlcolor{RGB|22,22,255}\textcolor{white}{\hl{év}}%
      \sethlcolor{white}\textcolor{black}{\hl{idents.}}%
    } \\

\multicolumn{4}{l}{\bfseries 6B-55B shared} \\
\midrule
$[0.56,\;0.27,\;0.17]$ & 15758 & eng & \makecell[l]{Detects head of multi-token or compound nouns} &
    \makecell[l]{%
      \sethlcolor{white}\textcolor{black}{\hl{The }}%
      \sethlcolor{RGB|22,22,255}\textcolor{white}{\hl{nan}}%
      \sethlcolor{RGB|100,100,255}\textcolor{white}{\hl{og}}%
      \sethlcolor{RGB|150,150,255}\textcolor{white}{\hl{ener}}%
      \sethlcolor{RGB|200,200,255}\textcolor{white}{\hl{ator}}%
      \sethlcolor{white}\textcolor{black}{\hl{ generates output ...}}%
    }
 \\
$[0.36,\;0.36,\;0.28]$ & 12525 & eng,fra,hin,por,spa & \makecell[l]{Boss concept detector\\[-0.7ex](\eg{}, \emph{chief}, \emph{jefe}, \emph{chefs}, \emph{chefe}, {\dn \7mHy} - boss)} &  
    \makecell[l]{%
      \sethlcolor{white}\textcolor{black}{\hl{O }}%
      \sethlcolor{RGB|22,22,255}\textcolor{white}{\hl{chefe}}%
      \sethlcolor{white}\textcolor{black}{\hl{ do departamento, Prof. ...}}%
    } \\
$[0.35,\;0.41,\;0.24]$ & 15248 & eng &\makecell[l]{Detects the token \emph{that} (only in English)} &  
    \makecell[l]{%
        \sethlcolor{white}\textcolor{black}{\hl{good number of analogies }}%
        \sethlcolor{RGB|22,22,255}\textcolor{white}{\hl{that}}%
        \sethlcolor{white}\textcolor{black}{\hl{ can apply to ...}}%
    }
 \\

\multicolumn{4}{l}{\bfseries 55B-341B shared} \\
\midrule
$[0.05,\;0.31,\;0.64]$ & 6997 & arb,eng,fra,hin,por,spa & \makecell[l]{Proper‑noun/ID detector that activates on named-entity heads} &  
    \makecell[l]{%
      \sethlcolor{RGB|150,150,255}\textcolor{white}{\hl{observers}}%
      \sethlcolor{white}\textcolor{black}{\hl{, the}}%
      \sethlcolor{RGB|22,22,255}\textcolor{white}{\hl{ June}}%
      \sethlcolor{white}\textcolor{black}{\hl{ 2015 }}%
      \sethlcolor{RGB|22,22,255}\textcolor{white}{\hl{King}}%
      \sethlcolor{RGB|150,150,255}\textcolor{white}{\hl{ v}}%
      \sethlcolor{white}\textcolor{black}{\hl{.}}%
      \sethlcolor{RGB|150,150,255}\textcolor{white}{\hl{ Bur}}%
      \sethlcolor{white}\textcolor{black}{\hl{well decision}}%
    }
 \\
    
\multicolumn{4}{l}{\bfseries 6B-55B-341B shared} \\
\midrule
$[0.35,\;0.32,\;0.32]$ & 12140 & arb,eng,fra,por,spa & \makecell[l]{Multilingual relative pronoun detector\\[-0.7ex](\eg{}, \emph{que}, \emph{that}, \emph{who}, \emph{aladhi})} &  
    \makecell[l]{%
  \sethlcolor{white}\textcolor{black}{\hl{predijo que los editores }}%
  \sethlcolor{RGB|10,10,255}\textcolor{white}{\hl{que}}%
  \sethlcolor{RGB|175,175,255}\textcolor{white}{\hl{ prosper}}%
  \sethlcolor{RGB|100,100,255}\textcolor{white}{\hl{arán}}%
  \sethlcolor{white}\textcolor{black}{\hl{ en el futuro}}%
}
  \\
$[0.32,\;0.31,\;0.37]$ & 4610 & eng,fra,por,spa & \makecell[l]{Phrasal‑verb/PP‑complement detector that fires on first\\[-0.7ex] token of verb‑plus‑particle or adjective‑plus‑preposition pattern} &  
    \makecell[l]{%
  \sethlcolor{white}\textcolor{black}{\hl{... army was}}%
  \sethlcolor{RGB|200,200,255}\textcolor{white}{\hl{ acc}}%
  \sethlcolor{RGB|150,150,255}\textcolor{white}{\hl{ustomed}}%
  \sethlcolor{RGB|22,22,255}\textcolor{white}{\hl{ to}}%
  \sethlcolor{white}\textcolor{black}{\hl{ in Europe. In the ...}}%
}
 \\
$[0.39,\;0.26,\;0.35]$ & 5819 & arb,eng,fra,spa & \makecell[l]{Activates most on new beginning of clauses\\[-0.7ex]right after a punctuation and wanes until a new clause} &  
\makecell[l]{%
  \sethlcolor{RGB|22,22,255}\textcolor{white}{\hl{some}}%
  \sethlcolor{RGB|100,100,255}\textcolor{white}{\hl{ readers}}%
  \sethlcolor{RGB|150,150,255}\textcolor{white}{\hl{ may}}%
  \sethlcolor{RGB|200,200,255}\textcolor{white}{\hl{ recall,}}%
  \sethlcolor{RGB|100,100,255}\textcolor{white}{\hl{ I}}
  \sethlcolor{RGB|200,200,255}\textcolor{white}{\hl{ cut}}
  \sethlcolor{white}\textcolor{black}{\hl{ my night photography}}%
}
 \\
\bottomrule
\end{tabular}
}%
\caption{\textbf{Subset of 3-way Crosscoder CLAMS French/English Annotation for BLOOM-1B  | 6B\compar{}55B \compar{}341B.} Languages are those that appeared when observing the feature's top-activating sentences. Early features are often language-specific, but over time these consolidate into crosslingual detectors capturing shared syntax and semantics.}
\label{tab:3way_annotation_bloom_clams_fraeng_subset}
\end{table*}

Fig.~\ref{fig:ie-evolution-pythia} tracks the top‑5 and bottom‑5 IE features from Pythia's early (1B) to final (286B) checkpoints. We observe a sharp decline in low‑level token detectors (\eg{}, subtokens or non‐interpretable features, depicted as those that do not have annotations) alongside a rise in grammatical detectors (\eg{}, prepositions, plural‑noun classes). A similar but longer-horizon trend appears in OLMo (Fig.~\ref{fig:ie-evolution-olmo}, Appendix~\ref{sec:appendix_analyses}), where for the rest of the training until 3T tokens, more abstract grammatical concept detectors emerge, such as those identifying plural nouns that depict jobs or skill attributes.

The manual annotations for Pythia-1B in Table~\ref{tab:3way_annotation_pythia_subset} illustrate this shift more in detail. Early checkpoints (1B-4B) employ detectors for specific tokens (\eg{}, the token \emph{-ans} in different contexts) or irregular forms (\eg{}, \emph{woman} vs.\ \emph{women}). By 4B, the model instead favors abstract, group‐level concepts, such as detecting multi-word nouns common in scientific writing or nouns denoting groups of humans. Between 4B and 286B, although overall performance plateaus, features with targeted linguistic functions---such as detectors for deverbal nominalizations---continue to emerge and strengthen (see Appendix~\ref{sec:appendix_annotations} for a complete list of list annotation). We additionally observe the same trend with the pairwise crosscoder comparisons (Table~\ref{tab:2way_annotation_pythia}), where the earlier checkpoint 128M employs more token-specific and edge-case detecting features while later checkpoints such as 4B develop detectors for functional token groups (\eg{}, plural quantifiers, such as \emph{many}, driving verb agreement).

\subsection{Feature Trajectories: Quantifying Emerging, Persistent, and Vanishing Causal Features via \relie{}}

To track how the causal importance of features redistributes across three checkpoints (\eg{}, 1B, 4B, and 286B), we conduct a three‑way \relie{} analysis (Eq.~\ref{eq:relie3way}), shown in Fig.~\ref{fig:monolingual-overlap}. Each axis corresponds to the one‑versus‑all \relie{} score for one checkpoint: a value of 1.0 means that a feature only plays a causal role for that checkpoint, while 0.0 means it has no effect on model behavior at that checkpoint. When we visualize the top-100\footnote{Given the $\min$ IE threshold 0.1, some tasks yield fewer than 100 features; in those cases, we use all that surpass it.} IE features using this method, several patterns emerge. 

High-importance features are primarily shared between 4B and 286B, consistent with their matching performance. However, many features also cluster near the 4B- and 286B-specific corners, indicating that even after performance plateaus, the model’s internal concepts continue to evolve and new features continue to arise. 

There is also substantial overlap across all three checkpoints and between 1B and 4B, where annotations indicate a shift from token-specific patterns to more abstract grammatical role detectors. Interestingly, despite being a sparser region, some features appear to be shared between 1B and 286B, at (0.5, 0.0, 0.5) coordinates. However, these shared features are generally less interpretable than those in other regions, often activating on sequences of random tokens, punctuation, or newlines. A version of the plot highlighting the top-10 IE features is provided in Appendix~\ref{sec:appendix_analyses}.

 \begin{figure*}[th]
    \centering
    \begin{subfigure}{0.28\linewidth}
      \centering
      \includegraphics[width=\textwidth]{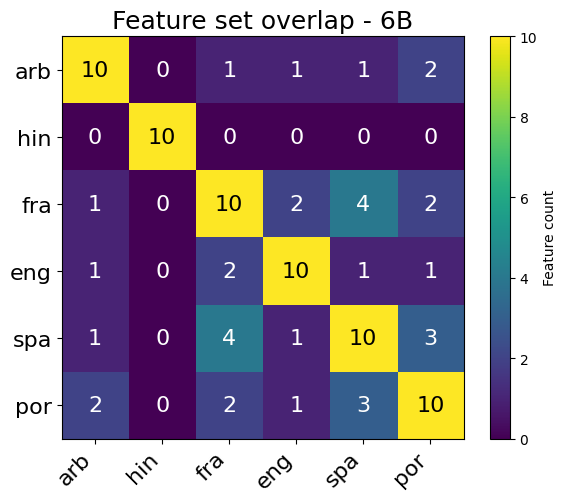}
    \end{subfigure}
    \begin{subfigure}{0.28\linewidth}
      \centering
      \includegraphics[width=\textwidth]{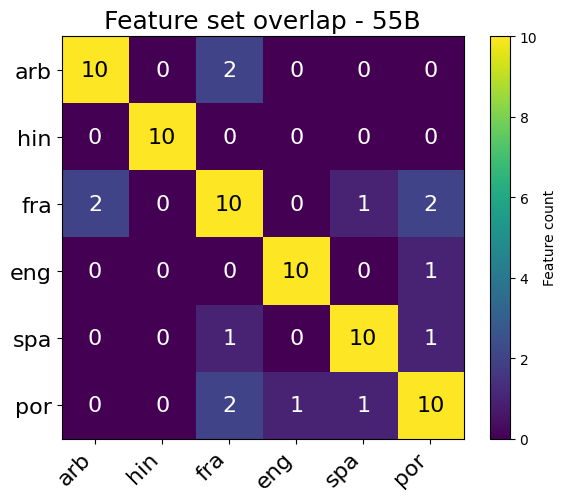}
    \end{subfigure}
    \begin{subfigure}{0.28\linewidth}
      \centering
      \includegraphics[width=\textwidth]{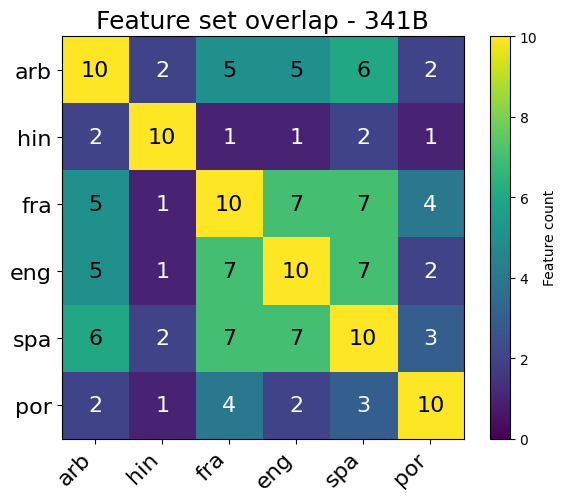}
    \end{subfigure}
    \caption{\textbf{Top-10 MultiBLiMP Number Agreement Task Feature Overlap per Checkpoint for Languages with 3-way comparisons.} Cross-lingual feature overlap increases by 341B, especially for Latin-script languages, reflecting shared syntactic patterns, while greater morphological complexity in Arabic and Hindi limits alignment.}
    \label{fig:multilingual-overlap-3way-subset}
    \vspace{-0.5em}
\end{figure*}

\section{Crosslingual Alignment of Syntactic Features in Multilingual Models}

Building on our findings in monolingual models---where lower‑level, token‑specific detectors give way to more abstract grammatical features---we now examine how multilingual LMs like BLOOM  learn to share features across languages.

\subsection{Consolidation of Monolingual Features into Multilingual Ones}
In Table~\ref{tab:3way_annotation_bloom_clams_fraeng_subset}, we show a subset of annotations for features significant for the CLAMS task in English and French. We observe that features of early checkpoints, (\eg{}, 6B) often have individual features for specific languages. For instance, instead of a single crosslingual feature that detects main-verb heads, the model maintains a separate detector for English. The crosslingual features found at this early stage are punctuation and delimeter detectors, which can be easier to abstract due to shared punctuation scripts. According to our pairwise checkpoint comparison, monolinguality of features in early phases also holds for conjunction and relative pronoun detecting features. As training progresses, these features merge into crosslingual ones. Checkpoints 6B and 55B, for example, share many such features among which we note the emergence of higher‑level semantic detectors, such as a crosslingual \emph{boss} concept feature, which likely reflects the prominence of that noun in our IE dataset. In the final stages we find crosslingual features for more complex constructions, such as one for detecting adjective-plus-preposition patterns.

\subsection{Quantifying Cross‑Lingual Feature Alignment and its Limits}

To quantify the number of crosslingual features over time, we calculate their overlap across five languages available in BLOOM and MutliBLiMP (Arabic, Hindi, English, French, Spanish, and Portuguese) and three subtasks (number, person, and gender agreement --- SV-\#, SV-P, and SV-G). In particular, we compute an IE score for each language–subtask pair and then measure how many of their top‑10 features intersect. We repeat this for each model checkpoint in our 3‑way comparison crosscoder (6B \compar{} 55B \compar{} 341B). Fig.~\ref{fig:multilingual-overlap-3way-subset} illustrates the SV-\# trend: first, a moderate overlap in the beginning of the training (6B) followed up by a mild overlap decrease for French, English, Spanish, and Portuguese at 55B (a pattern that we also observe for the remaining tasks in Fig.~\ref{fig:multilingual-overlap-3way} in Appendix~\ref{sec:appendix_analyses}). Despite the 55B checkpoint having similar performance on the task as 341B, the latter has a significantly higher number of overlapping features, especially for Latin script languages. We hypothesize two reasons: (1) the shared script leading to high overlap of token-specific features from the beginning of pretraining; and (2) having verbs that agree with their subjects in fairly predictable ways where nouns also often agree with adjectives in gender/number. On the other hand, Hindi and Arabic have more complex agreement systems (\eg{}, in Hindi, verbs agree with both subject and object in person/number/gender depending on verb aspect).

This raises an important question: how do LMs handle agreement in languages with greater morphological complexity? To unpack why Hindi has lower overlap, we examine the top IE‑annotated features per language in Appendix Tables~\ref{tab:3way_annotation_bloom_multiblimp_eng} (English),~\ref{tab:3way_annotation_bloom_multiblimp_fra} (French), and~\ref{tab:3way_annotation_bloom_multiblimp_hin} (Hindi). We find that high IE features for Hindi encode more information on the verbal aspect and the object than English and French. Importantly, we find that a majority of the overlap between Hindi and other languages is due to punctuation or parenthesis detecting low-level features. Our analysis shows that while multilingual models can learn a joint feature space for languages with similar morphological systems, languages with more complex or under‑represented agreement mechanisms---like Hindi and Arabic---may retain language‑specific representations even at larger scales for this particular middle layer in BLOOM. Future work should further examine whether and how such language-specific representations persist across layers. 
\section{Conclusion}
We deployed crosscoders to learn joint feature spaces between model checkpoints and introduced \relie{} to show when individual features become causally important for task performance. 
This lets us trace fine-grained changes in linguistic representations over pretraining, including which features emerge, persist, and disappear. 
We find that monolingual models transition from detecting specific tokens to high-level syntactic patterns, while multilingual models consolidate these into universal crosslingual features, reflecting increasingly shared representations.
This approach generalizes across architectures and scales to billion-parameter models. Future work could extend this analysis from individual features to circuits \citep{tigges2024circuit, hakimi-etal-2025-time}, tracing how distributed patterns of computation evolve over training and across layers, rather than relying solely on static slices of the model.

\section*{Acknowledgements}
We are grateful to Madhur Panwar, Negar Foroutan, Badr AlKhamissi, Gail Weiss, and Anja Surina for their helpful discussions and feedback on our manuscript. Antara Bhattacharya and Yugesh Kothari provided essential support in verifying the Hindi feature annotations. We also thank Julian Minder for insights on training crosscoders and Michael Hanna for the \LaTeX{} code used to visualize sequence activations in tables.

We also gratefully acknowledge the support of the Swiss National Science Foundation (No. 215390), the European Research Council (Starting grant no. 101222478, RESPECT-LM), the AI2050 program at Schmidt Sciences (Grant \#G-25-69783), Sony Group Corporation, and the Swiss National Supercomputing Center (CSCS) in the form of an infrastructure engineering and development project.

\section*{Limitations}
Our analysis is dependent on checkpoint selection, which can influence the takeaways drawn from the study. Early pretraining stage checkpoints are particularly hard to interpret, as features derived from them are generally less human-interpretable; thus, not all checkpoints will yield immediate insights with this setup. Additionally, our reliance on benchmarks like BLiMP may not fully capture the diversity of real-world linguistic variation; this limits the generalizability of our findings. 

The annotation process itself involves a degree of subjectivity, and our use of gradient attribution methods, such as integrated gradients, does not strictly guarantee causal relationships.\footnote{That said, past work has observed strong correlations between gradient attributions and exact interventions, especially when using integrated gradients \citep{marks2025featcircuits}} Finally, there is a risk of misinterpretation: it would be easy to falsely alias features into stable or human-like conceptual spaces, which could mislead downstream use or public understanding \citep{saphra2022creationism}. In the other direction, it is also not guaranteed that all atomic concepts used by the model are human understandable or explainable with natural language \cite{hewitt2025cantunderstandai}, similar to AlphaGo Zero's surprising ``nonstandard strategies beyond the scope of traditional Go knowledge'' \citep{silver2017masteringgowohuman}.

Finally, the computational cost of training crosscoders is non-trivial, although it is relatively short for typical SAE training due to our focus on the middle layer activations: we coarsely estimate 6 hours on an A100 80GB GPU for 2-way comparisons and 12 hours for 3-way comparisons in 1B models. Finally, scaling beyond 7B parameters across many checkpoint comparisons presents challenges, though this can potentially be mitigated by computing each source separately and aggregating the crosscoder latents iteratively.

\section*{Ethics Statement}
Our findings on crosslingual feature consolidation may help reveal language-specific underperformance and inequities, contributing to better understanding of multilingual model fairness. However, sharing detailed mappings of internal model representations carries a dual-use risk: while it can aid safety researchers, it may also enable malicious actors to design more sophisticated adversarial attacks or exploitations.

\bibliography{custom}

\appendix
\section{Preliminaries Continued}
\label{sec:appendix_premilinaries}

\paragraph{Sparse Autoencoders}
Sparse Autoencoders (SAE) learn to to reconstruct a model’s dense internal representation by projecting it onto a larger yet sparsely activating feature space with an encoder, and then decoding it back into the original activations. Formally, one way to implement SAEs is to enforce a sparsity and a reconstruction objective, such as the $\ell_1$ sparsity and $\ell_2$ reconstruction loss:
\begin{align} 
\fv &= \relu (W_{\enc} \xv + \bv_{\enc}) \\
\xhatv &= W_{\dec} \fv + \bv_{\dec} \\
\Loss &= \lVert \xv - \xhatv \rVert_{2}^{2} + \lambda \sum\limits_{i} \fv_i \lVert W_{\dec,i} \rVert_{2}
\end{align}
Beyond this implementation, there are a variety of alternative objectives. Some change how the ReLU activation function is applied (\eg{}, replacing it with JumpReLU; \citealp{rajamanoharan2024jumprelu}). Others modify the sparsity objective by directly applying a top-$k$ activation constraint instead of a open-ended $\ell_1$ sparsity loss \citep{makhzani2014ksae, bussmann2024batchtopksae, gao2025scaling}. Some also replace the direct reconstruction objective with a KL divergence loss between the model's original output distribution as the reference and its output when using the reconstructed activation in the forward pass \citep{braun2024e2e}.

\section{Dataset Details}
\label{sec:appendix_dataset}

\begin{table*}[ht]
\centering
\resizebox{1.0\linewidth}{!}{
\begin{tabular}{llp{8cm}ll}
\hline
\mythead{Dataset} & \mythead{Subtask} & \mythead{Prefix} & \mythead{Correct} & \mythead{Wrong} \\
\hline
\multirow{2}{*}{BLiMP} 
& Distractor relational noun & The granddaughters of every customer \underline{\hspace{0.5cm}} & don & doesn \\
& Irregular plural subject & The octopi \underline{\hspace{0.5cm}} & have & has \\
\hline
\multirow{2}{*}{MultiBLiMP} 
& Spanish & Desde algún lugar donde habita el recuerdo \underline{\hspace{0.5cm}} & fue & fui \\
& French & Cette proposition d'amendement a pour but que l'on tienne compte de les grands froids que \underline{\hspace{0.25cm}} & peuvent & pouvez \\
\hline
\multirow{2}{*}{CLAMS} 
& Simple Agreement & The farmer \underline{\hspace{0.5cm}} & is & are \\
& VP Coord & Les clients retournent et démén \underline{\hspace{0.5cm}} & agent & age \\
\hline
\end{tabular}
}
\caption{\textbf{Examples from subject--verb agreement tasks,} showing shared prefix with correct and wrong single-token completions. For BLiMP, we use the Pythia tokenizer, and for the other two multilingual datasets, we use BLOOM. BLiMP and CLAMS subtasks differ in terms of grammatical case and difficulty, while MultiBLiMP and CLAMS additionally include multilingual examples.}
\label{tab:subjectverb-examples}
\end{table*}

In this section we describe the datasets and the preprocessing steps. Note that we did not collect these datasets ourselves; they may contain personally identifying or offensive content.

\paragraph{Pile and DOLMA Subsampling}
To train crosscoders for Pythia and OLMo, we use the Pile and DOLMA datasets respectively \citep{gao2020pile, soldaini-etal-2024-dolma}. Specifically, we randomly subsample around 400M tokens from each dataset for training and around 120'000 tokens for the validation split. All reported metrics are calculated with the validation split.

\paragraph{mC4 Subsampling with ROOTS Ratios} 
We also subsample 400M tokens for BLOOM's crosscoder training dataset. However, because the original training dataset ROOTS is spread across multiple repositories on Huggingface Hub\footnote{\url{https://huggingface.co/bigscience-data/datasets}} \citep{laurencon2023roots}, we instead extract the same ten most-frequent languages from mc4 in identical proportions \citep{xue-etal-2021-mt5}: English (35\%), Chinese (19\%), French (15\%), Spanish (13\%), Portuguese (6\%), Arabic (5\%), Vietnamese (3\%), Hindi (2\%), Indonesian (1\%), and Bengali (1\%).

\paragraph{Subject--Verb Agreement Task Examples} 
We evaluate subject--verb agreement using three datasets: BLiMP \citep{warstadt-etal-2020-blimp-benchmark}, MultiBLiMP \citep{jumelet2025multiblimp}, and CLAMS \citep{mueller-etal-2020-cross}. Across all three benchmarks we apply a unified preprocessing pipeline: we identify a common prefix and isolate the tokens whose prediction reflects a correct versus incorrect agreement, which makes the process model-specific. We provide two examples from each task in Table~\ref{tab:subjectverb-examples}.

In BLiMP, we focus on four subtasks: (1) \texttt{Distractor agreement relational noun}, (2) \texttt{Distractor agreement relative clause},  (3) \texttt{Regular plural subject verb agreement 1}, (4) \texttt{Irregular plural subject verb agreement 1}. For the latter two subtasks, we omit their version 2 as they do not match our requirements for a single token completion given a single shared prefix. The processing yields a dataset containing 3577 entries for both Pythia and OLMo, with roughly similar amounts of examples from each subtask.

For MultiBLiMP, we restrict to the six languages that both appear in BLOOM’s top ten and have sufficient data (English, French, Spanish, Portuguese, Arabic, and Hindi). We use all valid examples when computing language-specific Indirect Effects (IEs): 575 examples for English, 1593 for French, 1242 for Spanish, 1481 for Portuguese, 692 for Arabic, and 964 for Hindi), but sample uniformly when we learn the IEs for several languages at once. When uniform sampling, each language has 100 examples from SV-\#, 100 examples from SV-G (except English, French, and Spanish, where no examples exist in MultiBLiMP), and 290 from SV-P.

For CLAMS, we only include the two overlapping languages with BLOOM, which are English and French, and we subsample from both using subtasks (1) \texttt{long\_vp\_coord} (300 for both), (2) \texttt{obj\_rel\_across\_anim} (400), (3) \texttt{obj\_rel\_within\_anim} (400), (4) \texttt{prep\_anim} (400), (5) \texttt{simple\_agrmt} (80), (6) \texttt{subj\_rel} (400), and (7) \texttt{vp\_coord} (400), resulting in a total of 4760 examples.

\section{Model Details} 
\label{sec:appendix_model}

All experiments use the 1B-parameter variants of Pythia, OLMo(1), and BLOOM. We choose to use OLMo(1)-1B instead of OLMo2-1B \citep{olmo20252olmo2furious} as the former has more frequent checkpointing during pretraining. These models differ in depth and hidden dimension size, as shown in Table~\ref{tab:specific-hyperparams}. Pythia and OLMO's 1B version have 16 layers, whereas BLOOM has 24, hence why we learn a crosscoder at the 8th layer for Pythia and OLMo, and at the 12th layer for BLOOM.

\begin{table}[t]
\centering
\resizebox{0.7\linewidth}{!}{
\begin{tabular}{rc}
\toprule
\mythead{Hyperparameter} & \mythead{Value} \\
\midrule
\texttt{seeds} & [124, 153, 6582]  \\ \midrule
\texttt{num\_train\_tokens} & 400M \\
\texttt{train\_batch\_token\_num} & 4096 \\
\texttt{val\_batch\_token\_num} & 8184 \\
\texttt{num\_val\_batches} & 30 \\ \midrule
\texttt{dict\_size} & 16384 \\
\texttt{dec\_init\_norm} & 0.08 \\
\texttt{enc\_dtype} & fp32 \\ \midrule
\texttt{lr} & 5e-05 \\
\texttt{l1\_warmup\_pct} & 0.05 \\
\texttt{l1\_coeff} & 2 \\
\texttt{beta1} & 0.9 \\
\texttt{beta2} & 0.999 \\
\bottomrule
\end{tabular}
}
\caption{\textbf{Hyperparameters for Crosscoder Training.}}
\label{tab:common-hyperparams}
\end{table}
\begin{table}[t]
\centering
\resizebox{1.0\linewidth}{!}{
\begin{tabular}{rcc}
\toprule
\mythead{Hyperparameter} & \mythead{Pythia-1B and OLMo-1B} & \mythead{BLOOM-1B} \\
\midrule
\texttt{hidden\_dim} & 2048 & 1536 \\
\texttt{num\_layers} & 16 & 24 \\
\texttt{mid\_layer} & 8 & 12 \\
\bottomrule
\end{tabular}
}
\caption{\textbf{Model-specific Hyperparameters.} Note that the HF model names we used to load these models are \texttt{pythia-1b}, \texttt{OLMo-1B-0724-hf}, \texttt{bloom-1b1-intermediate}.}
\label{tab:specific-hyperparams}
\end{table}

They also differ in their tokenizers, position embeddings, attention implementation, and activation functions, which can affect how models process their mid-layer output. All three models use a BPE style tokenizer \citep{sennrich-etal-2016-neural}. In addition, OLMo and Pythia uses RoPE embeddings \citep{su2024rope}, while BLOOM uses ALiBi \citep{press2022alibi}. BLOOM also implements a multi-query attention where each head has its own query but shares key and value projections, whereas the other models have separate key/value/query per heads. Finally Pythia and BLOOM use a GeLU activation function \citep{hendrycks2023gelu}, whereas OLMo uses SwiGLU \citep{shazeer2020gluvariantsimprovetransformer}.

\section{Crosscoder Training Details}
\label{sec:appendix_crosscoders}

\paragraph{Hardware}
For training we use a single 80GB NVIDIA A100 GPU. Under our chosen hyperparameters, pairwise (2‑way) comparisons converge in about 6 hours, while 3‑way experiments require closer to 12. By contrast, training sparse autoencoders for several layers can stretch into multiple days. Because our batch sizes, learning rates, and model dimensions fit comfortably within one 80GB GPU’s memory, we did not pursue any data or model parallelization schemes.

\paragraph{Hyperparameters}
We share our hyperparameters in Tables~\ref{tab:common-hyperparams} and~\ref{tab:specific-hyperparams}. All crosscoders are trained using three fixed random seeds, $\{124,153,6582\}$. Training proceeds until approximately $400\mathrm{M}$ tokens have been processed. Optimization is performed with Adam, using a peak learning rate of $5\times10^{-5}$, $\beta_1=0.9$, and $\beta_2=0.999$. The $\ell_1$ penalty on decoder‐weighted activations is ramped up linearly over the first $5\%$ of training, reaching a coefficient of $\lambda=2$. Decoder weights are randomly initialized with norm $0.08$, and we employ a dictionary size of $16\,384$. 

\begin{table}[t]
  \centering
  \resizebox{1.0\linewidth}{!}{
    \begin{tabular}{ 
      cc 
      @{\hspace{2em}}
      *{5}{c} 
    }
      \toprule
      \multicolumn{2}{c}{} 
        & \multicolumn{5}{c}{\bfseries $\bm{\ell_1}$ Sparsity Crosscoder} \\
      \cmidrule(r){3-7}
      \textbf{Model} & \textbf{Comparison}
        & \textbf{$\bm{\ell_0}$} & \textbf{Dead Feats} 
        & \textbf{$\Delta$CE A} & \textbf{$\Delta$CE B}  & \textbf{$\Delta$CE C} \\
      \midrule
                    & 128M \compar{} 1B & 88 & 9 & 0.00 & 0.07 & - \\
        Pythia-1B   & 1B \compar{} 4B & 214 & 1 & 0.05 & 0.18 & - \\
        Layer 8     & 4B \compar{} 286B & 190 & 9 & 0.15 & 0.48 & - \\
                    & 1B \compar{} 4B \compar{} 286B & 215 & 19 & 0.03 & 0.16 & 0.54 \\
      \midrule
                    & 2B \compar{} 4B & 184 & 0 & 0.08 & 0.20 & - \\
        OLMo-1B     & 4B \compar{} 33B & 227 & 0 & 0.14 & 0.21 & - \\
        Layer 8     & 33B \compar{} 3048B & 182 & 425 & 0.16 & 0.35 & - \\
                    & 4B \compar{} 33B \compar{} 3048B & 225 & 101 & 0.12 & 0.18 & 0.43 \\
      \midrule
                    & 550M \compar{} 6B & 211 & 6 & 0.10 & 0.20 & - \\
        BLOOM-1B    & 6B \compar{} 55B & 112 & 8 & 0.14 & 0.29 & - \\
        Layer 12    & 55B \compar{} 341B & 96 & 12 & 0.18 & 0.18 & - \\
                    & 6B \compar{} 55B \compar{} 341B & 118 & 19 & 0.13 & 0.20 & 0.22 \\
    \bottomrule
    \end{tabular}
  }
  \caption{\textbf{Crosscoder statistics.} Results averaged over three seeds on validation set. $\Delta$CE is the change in cross-entropy loss when doing a forward pass using the original output versus the crosscoder reconstruction. A, B, C refer to the 1st, 2nd and 3rd checkpoints used for loss computation. $\ell_0$ and dead feature averages are rounded to integers. Less trained models (\eg{}, 1B) get smaller $\Delta$CE values than further trained models (\eg{}, 286B) due to the former's high original CE loss.}
  \label{tab:crosscoder-stats}
\end{table}

\begin{table*}[ht]
\centering
\small
\resizebox{1.0\linewidth}{!}{
\begin{tabular}{l
    lrr
    @{\hspace{2em}}
    lrr}
\toprule
\multicolumn{1}{c}{\multirow{2}{*}{\textbf{Task}}}
& \multicolumn{3}{c@{\hspace{2em}}}{\textbf{Pythia-1B}} 
& \multicolumn{3}{c}{\textbf{OLMo-1B}} \\
\cmidrule{2-4} \cmidrule{5-7}
& \textbf{Comparison} & \(\rho(\frac{|\Delta c_2|}{|\Delta c_1|},\mathrm{RelDec})\) & \(\rho(\frac{|\Delta c_2|}{|\Delta c_1|},\mathrm{RelIE})\)
& \textbf{Comparison} & \(\rho(\frac{|\Delta c_2|}{|\Delta c_1|},\mathrm{RelDec})\) & \(\rho(\frac{|\Delta c_2|}{|\Delta c_1|},\mathrm{RelIE})\) \\
\midrule
\texttt{Distractor Relational Noun} 
& 128M \compar{} 1B & 0.316 & \textbf{0.934} 
& 2B \compar{} 4B & 0.920 & \textbf{0.972} \\
& 1B \compar{} 4B & 0.930 & \textbf{0.958} 
& 4B \compar{} 33B & \textbf{0.949} & 0.897 \\
& 4B \compar{} 286B & 0.691 & \textbf{0.964} 
& 33B \compar{} 3048B & 0.770 & \textbf{0.973} \\
\midrule
\texttt{Distractor Relative Clause} 
& 128M \compar{} 1B & 0.788 & \textbf{0.966} 
& 2B \compar{} 4B & 0.961 & \textbf{0.979} \\
& 1B \compar{} 4B & 0.901 & \textbf{0.922} 
& 4B \compar{} 33B & \textbf{0.956} & 0.938 \\
& 4B \compar{} 286B & 0.784 & \textbf{0.941} 
& 33B \compar{} 3048B & 0.771 & \textbf{0.989} \\
\midrule
\texttt{Irregular Plural Subject} 
& 128M \compar{} 1B & 0.941 & \textbf{0.979} 
& 2B \compar{} 4B & \textbf{0.982} & 0.966 \\
& 1B \compar{} 4B & 0.777 & \textbf{0.937} 
& 4B \compar{} 33B & 0.898 & \textbf{0.905} \\
& 4B \compar{} 286B & 0.843 & \textbf{0.954} 
& 33B \compar{} 3048B & 0.785 & \textbf{0.810} \\
\midrule
\texttt{Regular Plural Subject} 
& 128M \compar{} 1B & 0.794 & \textbf{0.948} 
& 2B \compar{} 4B & 0.838 & \textbf{0.966} \\
& 1B \compar{} 4B & 0.874 & \textbf{0.930} 
& 4B \compar{} 33B & \textbf{0.961} & 0.919 \\
& 4B \compar{} 286B & 0.806 & \textbf{0.908} 
& 33B \compar{} 3048B & 0.862 & \textbf{0.956} \\
\midrule
\textbf{Avg by Comparison}
& Avg 128M \compar{} 1B & 0.710 & \textbf{0.957} 
& Avg 2B \compar{} 4B & 0.925 & \textbf{0.971} \\
& Avg 1B \compar{} 4B & 0.870 & \textbf{0.937} 
& Avg 4B \compar{} 33B & \textbf{0.941} & 0.915 \\
& Avg 4B \compar{} 286B & 0.781 & \textbf{0.942} 
& Avg 33B \compar{} 3048B & 0.797 & \textbf{0.932} \\
\midrule
\textbf{Overall Avg} 
& - & 0.787 & \textbf{0.945} 
& - & 0.843 & \textbf{0.952} \\
\bottomrule
\end{tabular}
}
\caption{\textbf{Top-10 significant feature ablation for Pythia-1B and OLMo-1B.} Spearman correlations \(\rho\) between the ratio of log-probability-difference metrics and model-attributing scores (\reldec{}, \relie{}), across four subject–verb agreement phenomena and various phase transition comparisons. \relie{} shows consistently higher correlations than \reldec{} across tasks and model comparisons, indicating that focusing on task-relevant signal uncovers more meaningful and stable task-specific features.}
\label{tab:corr_top10_combined}
\vspace{-1em}
\end{table*}

\paragraph{Crosscoder Learning Results}
In Table~\ref{tab:crosscoder-stats} we show the averaged metrics of trained crosscoders across three seeds. $\ell_0$ records how many crosscoder features fire on average per token. Dead features refer to the number of features that were never activated across all validation batches. $\Delta$CE shows the difference in cross-entropy loss when doing a forward pass using the original mid-layer output versus the crosscoder reconstruction. A, B, C refer to the first, second and third checkpoints used for loss computation. For a detailed analysis, see \S\ref{subsec:learnability} in the main paper.

\section{Attribution Correlation via Feature Ablation}
\label{sec:appendix_ablation}

To assess whether \relie{}’s focus on task‑relevant signals yields a more targeted identification of significant features than \reldec{}, we perform an ablation study on the top-10 features deemed most important by IE on each BLiMP subtask (see Appendix~\ref{sec:appendix_dataset}). For each checkpoint, we measured the change in log probability difference upon ablating each feature (one-by-one), denoted \(\Delta c\), and then computed the ratio \(\lvert \Delta c_2 \rvert / \lvert \Delta c_1 \rvert\) to quantify which checkpoint’s predictions were more adversely affected. Given that 1.0 means more relevant for $c_2$ and 0.0 for $c_1$ for both \relie{} and \reldec{}, we expect a high positive correlation between the two if the feature was indeed more important for one checkpoint versus the other.

Table~\ref{tab:corr_top10_combined} presents Spearman correlations ($\rho$) between the ratios of log-probability-difference-differences $|\Delta c_2| / |\Delta c_1|$ and two checkpoint-attributing scores---\reldec{} and \relie{}---across four BLiMP subtasks. For both Pythia-1B and OLMo-1B, \relie{} has higher correlation with the ratio (avg $\rho$ = 0.945 and 0.952, respectively). In short, by focusing on task‐relevant signal rather than the entire feature set, \relie{} uncovers more interpretable and task-relevant features more effectively.

\section{Feature Annotations}
\label{sec:appendix_annotations}

\subsection{Indirect Effect Implementation}
\label{sec:appendix_ie}

As mentioned in \S\ref{sec:preliminaries}, we approximate the Indirect Effect (IE) of each crosscoder feature using integrated gradients (IG; \citealp{sundararajan2017axiomaticattr, hanna2024faith, marks2025featcircuits}). We use integrated gradients because they provide a principled attribution method with desirable axiomatic guarantees \citep{sundararajan2017axiomaticattr}, and prior work suggests that IG-based estimates of IE better match ablation-based measurements than alternative attribution methods \citep{kramar2024atpstar,marks2025featcircuits}. Formally, we approximate IE as:
\begin{equation}
\begin{split}
\ieig&(m; \mathbf{a}; x)
= (\mathbf{a}_{\text{patch}} - \mathbf{a}) \\
& \quad \times \frac{1}{N} \sum_{\alpha}
\left.\nabla_{\mathbf{a}} m\right|_{\alpha \mathbf{a}_{\text{clean}} + (1 - \alpha) \mathbf{a}_{\text{patch}}}
\end{split}
\end{equation}
where the sum ranges over $N=10$ equally-spaced $\alpha\in \{0,\tfrac{1}{N},\dots,\tfrac{N-1}{N}\}$ steps. This gradient-based formulation allows us to estimate the IE for all features simultaneously using only a fixed number of passes, rather than scoring each feature via a separate intervention. Thus, it is an $O(1)$ algorithm rather than $O(n)$, where $n$ is the number of features. After summing the gradients, we average the resulting IE scores across the batch (following \citealp{marks2025featcircuits}). Finally, we threshold the batch‑averaged IE values, setting any below 0.1 to zero.

\subsection{Annotation Instructions}
Annotations were done by the authors. For languages that none of the authors spoke natively, the annotation was completed with the use of translators and verified with at least one native speaker. In particular, during the annotation process, the expert was not exposed to the \relie{} values of the features; rather, they were asked to answer 4 questions adapted from \citeauthor{marks2025featcircuits}:
\begin{enumerate}[leftmargin=1.2em, itemsep=0pt, topsep=3pt]
    \item \textbf{Description:} To the best of your extent, describe the behavior of this feature's activation. 
    \item \textbf{Interpretability:} On a scale of 0.0 to 1.0, how coherent are the examples shown with the description you wrote? Is it consistently activating on similar tokens or promoting/demoting similar tokens?
    \item \textbf{Complexity:} On a scale of 0.0 to 1.0, how complex is the feature behavior? How broad is the topic that the feature fires on? Does the feature activate on or promote/demote diverse tokens or similar tokens all over again?
    \item (if BLOOM) \textbf{Languages:} Which languages have this feature activated most on?
\end{enumerate}

\subsection{Complete Annotations}

We provide here the full set of annotation tables omitted from the main paper: Pythia’s two‑way and three‑way comparisons (Tables~\ref{tab:2way_annotation_pythia} and~\ref{tab:3way_annotation_pythia}), OLMo’s three‑way comparison (Table~\ref{tab:3way_annotation_olmo}), and BLOOM’s two‑way and three‑way comparisons (Tables~\ref{tab:2way_annotation_bloom_clams_fraeng} and~\ref{tab:3way_annotation_bloom_clams_fraeng}). Language‑specific annotations for English, French, and Hindi appear in Tables~\ref{tab:3way_annotation_bloom_multiblimp_eng},~\ref{tab:3way_annotation_bloom_multiblimp_fra} and~\ref{tab:3way_annotation_bloom_multiblimp_hin}, respectively. We transliterated the Arabic text instead of rendering it in its native script, because the PDF \LaTeX{} compiler could not render Devanagari and Arabic script simultaneously.

\begin{table*}[!ht]
\centering
\resizebox{0.79\linewidth}{!}{%
\begin{tabular}{crl}
\toprule
\mythead{RelIE} & \mythead{FeatID} & \mythead{Interpreted Function} \\
\midrule
\multicolumn{3}{l}{\bfseries Comparison: 128M \compar{} 1B} \\
\midrule
\multicolumn{3}{l}{\itshape 128M specific} \\
\midrule
$0.19$ & 5667 & - \\
$0.28$ & 14250 & Detects token \emph{-ese} at the end of a word \\
$0.29$ & 440 & Detects token \emph{-ara} at the end of a name, promotes possession or verbs \\
\multicolumn{3}{l}{\itshape 128M-1B shared} \\
\midrule
$0.38$ & 8636 & - \\
$0.43$ & 3164 & Detects \emph{-us} ending, often for a Latin origin single noun and promotes verb \emph{is}  \\
$0.52$ & 12683 & Detects the noun \emph{analysis}  \\
$0.53$ & 1749 & Detects irregular plural noun \emph{people} \\
$0.56$ & 5032 & Detects irregular plural noun \emph{men}, promotes EOS, conjunction, or verbs \\
$0.57$ & 7072 & - \\
\multicolumn{3}{l}{\itshape 1B specific} \\
\midrule
$0.71$ & 15882 & Detects singular \emph{man}, promotes preposition completion \\
$0.83$ & 4118 & Detects singular \emph{woman}, not necessarily as a subject \\
$0.89$ & 6381 & - \\
$0.91$ & 10069 & Detects nouns that end with \emph{-ists} and promotes plural verb completion or prepositions \\
$0.92$ & 14897 & Detects nouns that end with \emph{-ans} \\
$0.94$ & 16118 & Detects words ending with \emph{-ias} \\
$0.95$ & 8757 & - \\
$1.00$ & 3811 & Detects regular plural nouns and promotes conjunction or prepositions \\
$1.00$ & 7483 & Detects singular nouns preceded by \emph{this} \\
\addlinespace
\multicolumn{3}{l}{\bfseries Comparison: 1B \compar{} 4B} \\
\midrule
\multicolumn{3}{l}{\itshape 1B specific} \\
\midrule
$0.00$ & 12677 & Detects regular plural nouns that refer to groups of people and promotes plural verb completion or prepositions \\
$0.10$ & 5778 & Detects \emph{man} starting words but promotes multi-token word completions (\eg{}, \emph{-hood}, \emph{-ned}, \emph{-hattan}) \\
$0.17$ & 1440 & Detects words containing the mid-token \emph{-es} and promotes medical term completions (\eg{}, \emph{-es-ophagus}) \\
$0.28$ & 3737 & Detects singular \emph{woman}, not necessarily as a subject \\
\multicolumn{3}{l}{\itshape 1B-4B shared} \\
\midrule
$0.37$ & 12685 & Detects nouns that end with \emph{-ans} \\
$0.48$ & 7616 & Detects nouns that end with \emph{-ists} and promotes plural verb completion or prepositions \\
$0.53$ & 14814 & Detects singular nouns preceded by \emph{this} in front of them, and promotes singular verb conjugation \\
$0.68$ & 11799 & Detects regular plural nouns that refer to objects/science concepts and promotes plural verb completion or prepositions \\
\multicolumn{3}{l}{\itshape 4B specific} \\
\midrule
$0.82$ & 9385 & - \\
$0.85$ & 744 & - \\
$0.88$ & 14210 & Detecs regular plural nouns \\
$1.00$ & 13102 & - \\
$1.00$ & 1868 & Detects punctuations and newline to promote BOS words \\
$1.00$ & 4050 & Detects nouns that are preceded by plural quantifiers (\eg{}, \emph{most}, \emph{many}, \emph{majority of}, \emph{some}) \\
$1.00$ & 9326 & Detects plural regular nouns, promotes plural conjugated verbs \\
$1.00$ & 9414 & Comma detector \\
$1.00$ & 11088 & Detects the final token of first names \\
$1.00$ & 14546 & Detects HTML/code-related regular plural object nouns \\
\addlinespace
\multicolumn{3}{l}{\bfseries Comparison: 4B \compar{} 286B} \\
\midrule
\multicolumn{3}{l}{\itshape 4B specific} \\
\midrule
$0.00$ & 4368 & - \\
$0.00$ & 2307 & Detects regular plural nouns that refer to science concepts and promotes plural verb completion or prepositions\\
$0.14$ & 479 & - \\
$0.19$ & 680 & - \\
$0.27$ & 13452 & Detects (regular and irregular) plural nouns, promotes plural verb completion \\
$0.27$ & 10514 & Detects regular plural nouns that can be followed up with \emph{themselves} \\
\multicolumn{3}{l}{\itshape 4B-286B shared} \\
\midrule
$0.37$ & 15084 & Detects (regular and irregular) plural nouns that refer to groups of people, promotes plural verb completion \\
$0.56$ & 5268 & - \\
$0.58$ & 5129 & - \\
\multicolumn{3}{l}{\itshape 286B specific} \\
\midrule
$0.79$ & 14815 & - \\
$0.85$ & 6511 & - \\
$0.89$ & 5588 & Detects newlines \\
$0.92$ & 12003 & - \\
$0.98$ & 13244 & Detects first names that are not followed up a last name \\
$0.98$ & 12108 & Detects HTML/code-related plural object nouns \\
$1.00$ & 2139 & Detects a larger variety of prepositions and complementizers (\eg{}, \emph{by}, \emph{from}, \emph{due}, \emph{with}, \emph{concerning}) \\
$1.00$ & 5138 & Detects last token of multi-token first names, promotes last names \\
\addlinespace
\bottomrule
\end{tabular}
}%
\caption{\textbf{2‑way L1‑Sparsity Crosscoder Annotation for Pythia-1B.} Each block is one pairwise comparison. \relie{} is sorted from 0.00 to 1.00, where < 0.3 gets attributed to the first checkpoint; > 0.7 to second; shared otherwise). Interpreted function gives a description if a linguistic role was detected, ``--'' otherwise. Pairwise comparisons reveal finer-grained feature shifts from one checkpoint to another, but cannot assess persistence like triplet analyses. This shows that early checkpoints (\eg{}, 128M) capture low-level lexical and morphological patterns, while slightly further trained ones (\eg{}, 1B) detect slightly more abstract patterns, such as irregular plurals.}
\label{tab:2way_annotation_pythia}
\end{table*}
\begin{table*}[!ht]
\centering
\resizebox{1.0\linewidth}{!}{%
\begin{tabular}{crp{18cm}l}
\toprule
\mythead{RelIE} & \mythead{FeatID} & \mythead{Interpreted Function} & \mythead{Languages} \\
\midrule
\multicolumn{4}{l}{\bfseries Comparison: 550M \compar{} 6B} \\
\midrule
\multicolumn{4}{l}{\itshape 550M specific} \\
\midrule
$0.00$ & 8760 & Detects administrative/government-related nouns & fra \\
$0.00$ & 14133 & Detects nouns and verbs that convey key actions, entities, or ideas in a sentence & eng \\
$0.05$ & 12275 & Detects conjunction token \emph{et} & fra \\
$0.10$ & 8341 & Detects subtoken \emph{et} typically conjunction but also for \emph{et al.} & fra \\
$0.20$ & 15852 & Detects head nouns and their modifiers that signal prominent participants or components & fra \\
$0.22$ & 14697 & Detects plural nouns, promotes plural verb conjugation and \emph{who} pronoun & eng,fra,spa \\
\multicolumn{4}{l}{\itshape 550M-6B shared} \\
\midrule
$0.36$ & 1474 & Detects plural French articles (\eg{}, \emph{les}, \emph{nos}, \emph{certains}), promotes plural single token noun completions & fra \\
$0.39$ & 8223 & Promotes \emph{-age} completion for nouns & eng,fra,spa \\
$0.44$ & 7882 & Detects English relative pronoun \emph{that}, promotoes pronoun follow ups & eng \\
$0.69$ & 14645 & Detects \emph{Ev} at BOS, to be completed with French adverbs or nouns & fra \\
\multicolumn{4}{l}{\itshape 6B specific} \\
\midrule
$0.95$ & 12523 & - & - \\
$0.96$ & 10853 & Detects noun and nominal expressions representing abstract entities, events, or processes & eng,fra,por,spa \\
$0.97$ & 2189 & Detects multi-word nouns (\eg{}, compound nouns or with adjectives) & eng,fra,por,spa \\
$1.00$ & 9386 & - & - \\
$1.00$ & 9813 & \emph{who} and \emph{that} detector & arb,eng,fra,por,spa \\
$1.00$ & 10337 & Punctuation and newline detector & arb,eng,fra,por,spa \\
$1.00$ & 14067 & - & - \\
$1.00$ & 15428 & Detects verbs with the concept to like/love/appreciate & eng,fra,hin*,por,spa \\
\addlinespace
\multicolumn{4}{l}{\bfseries Comparison: 6B \compar{} 55B} \\
\midrule
\multicolumn{4}{l}{\itshape 6B specific} \\
\midrule
$0.00$ & 11469 & Detects punctuation and newline & eng,fra,spa \\
$0.17$ & 942 & Verbs that depict dynamic, agentive actions & eng \\
$0.20$ & 10311 & Detects \emph{ev} or \emph{Ev} tokens to be completed with french adverbs or nouns & fra \\
$0.28$ & 15632 & Detects verbs with the concept to like/love/appreciate & arb*,eng,fra,por,spa \\
\multicolumn{4}{l}{\itshape 6B-55B shared} \\
\midrule
$0.31$ & 11920 & Detects head of noun phrases denoting concrete or informational entities (\eg{},\emph{data}, \emph{system}, \emph{text}) & eng,fra,por,spa \\
$0.32$ & 12000 & Detects noun that depict occupational or social roles like researchers, engineers, physician, and journalists & arb*,eng,fra,hin*,por,spa \\
$0.32$ & 2792 & - & - \\
$0.38$ & 14748 & - & - \\
$0.40$ & 5763 & Detects the concept boss in different languages (\eg{}, \emph{chef}, \emph{boss}, \emph{jefe}) probably because it is a common noun in the IE dataset & eng,fra,spa \\
$0.56$ & 9817 & Detects verbs and promotes preposition/conjunction/punctuation & eng,fra,por,spa \\
$0.56$ & 425 & Detects relative pronoun and prepositions (\eg{}, \emph{that}, \emph{que}, \emph{at}, \emph{of}, \emph{de}) & eng,fra,spa \\
$0.59$ & 4863 & Detects relative pronoun \emph{that} and promotes verb/pronoun completions & eng \\
\multicolumn{4}{l}{\itshape 55B specific} \\
\midrule
$0.79$ & 5345 & Nouns and verbs related to consultants and consulting & arb*,eng,fra,por,spa \\
\addlinespace
\multicolumn{4}{l}{\bfseries Comparison: 55B \compar{} 341B} \\
\midrule
\multicolumn{4}{l}{\itshape 55B specific} \\
\midrule
$0.27$ & 15249 & Nouns and verbs related to consultants and consulting & arb*,eng,fra,por,spa \\
$0.27$ & 12734 & - & - \\
\multicolumn{4}{l}{\itshape 55B-341B shared} \\
\midrule
$0.35$ & 11458 & Detects \emph{that} and promotes verb or pronoun completion in English & eng \\
$0.39$ & 11920 & Detects nouns to be completed by \emph{de}, \emph{of} & eng,fra,por,spa \\
$0.39$ & 7339 & Determiners and quantifiers in noun phrases, such as articles (\eg{}, \emph{a}, \emph{os}), possessives (\eg{} our), and universal quantifiers (\eg{} every, todo, cada) & eng,por,spa \\
$0.44$ & 6063 & Detects the concept boss in different languages (\eg{}, \emph{chef}, \emph{boss}, \emph{jefe}) probably because it is a common noun in the IE dataset and promotes \emph{of} (\eg{}, \emph{de}, \emph{do}, \emph{du}) completions & eng,fra,por,spa \\
$0.48$ & 15083 & Relative pronouns and the syntactic material inside relative clauses & eng,fra,por,spa \\
$0.54$ & 10325 & Detects \emph{ev} or \emph{Ev} tokens to be completed with french adverbs or nouns & fra,por,spa \\
$0.55$ & 425 & Detects relative pronouns/subordinators (\eg{}, \emph{that}, \emph{que}, \emph{qui}, \emph{which}, \emph{who}, \emph{où}) to introduce a new clause; also activates on the verbs inside the subordinate clause & arb*,eng,fra,por,spa \\
$0.57$ & 8729 & - & - \\
\multicolumn{4}{l}{\itshape 341B specific} \\
\midrule
$1.00$ & 794 & - & - \\
$1.00$ & 7419 & Detects newlines in different languages & arb,fra,hin,por,spa \\
$1.00$ & 7806 & - & - \\
$1.00$ & 13276 & - & - \\
$1.00$ & 13404 & Sentence-boundary detector through punctuation and other delimiters & arb,fra,por,spa,zh \\
\addlinespace
\bottomrule
\end{tabular}
}%
\caption{\textbf{2-way L1-Sparsity Crosscoder CLAMS French/English Annotation for BLOOM-1B.} Each block is one pairwise comparison. \relie{} is sorted from 0.00 to 1.00, where < 0.3 gets attributed to the first checkpoint; > 0.7 to second; shared otherwise). Interpreted function gives a description if a linguistic role was detected, ``--'' otherwise. Languages lists which languages the feature highly activates on, * means that the activation was relatively less common. While earlier checkpoints (\eg{}, 550M) capture language specific low-level function words, later checkpoints (\eg{}, 55B and 341B) increasingly share such features across languages.}
\label{tab:2way_annotation_bloom_clams_fraeng}
\end{table*}
\begin{table*}[!ht]
\centering
\resizebox{1.0\linewidth}{!}{%
\begin{tabular}{crl}
\toprule
\mythead{RelIE} & \mythead{FeatID} & \mythead{Interpreted Function} \\
\midrule
\multicolumn{3}{l}{\bfseries 1B-4B shared} \\
\midrule
$[0.53,\;0.33,\;0.15]$ & 1067 & Detects subtoken \emph{-ans} typically in names \\
$[0.41,\;0.39,\;0.20]$ & 941 & Detects plural nouns that are art-related professions \\
$[0.45,\;0.41,\;0.14]$ & 4897 & Detects plural nouns that end with \emph{-ists}, (\eg{}, \emph{protagonist}, \emph{capitalist}, \emph{pharmacist}) \\
$[0.32,\;0.43,\;0.25]$ & 15204 & Detects singular nouns found in technical discourse (\eg{}, \emph{method}, \emph{function}, \emph{guide}, \emph{recipe}) preceeded by the word "This" \\
\multicolumn{3}{l}{\bfseries 1B-286B shared} \\
\midrule
$[0.55,\;0.10,\;0.34]$ & 7489 & Detects singular \emph{woman} noun \\
$[0.52,\;0.03,\;0.45]$ & 1641 & Detects newlines \\
$[0.52,\;0.01,\;0.46]$ & 3852 & Detects singular \emph{man} noun \\
\multicolumn{3}{l}{\bfseries 4B specific} \\
\midrule
$[0.00,\;1.00,\;0.00]$ & 15556 & Detects a full stop and promotes connection words or newlines \\
$[0.00,\;1.00,\;0.00]$ & 11274 & Multi-word noun or compound noun detector \\
$[0.00,\;0.99,\;0.01]$ & 8318 & Detects regular plural nouns \\
$[0.00,\;0.96,\;0.04]$ & 10020 & - \\
$[0.11,\;0.69,\;0.20]$ & 15950 & Detects regular plural nouns \\
$[0.02,\;0.68,\;0.30]$ & 10523 & Detects plural nouns mostly depicting humans (\eg{}, \emph{people}, \emph{students}, \emph{bloggers}) \\
$[0.11,\;0.62,\;0.26]$ & 15118 & - \\
\multicolumn{3}{l}{\bfseries 4B-286B shared} \\
\midrule
$[0.01,\;0.30,\;0.69]$ & 11987 & - \\
\multicolumn{3}{l}{\bfseries 286B specific} \\
\midrule
$[0.00,\;0.00,\;1.00]$ & 15323 & Detects plural nouns found in technical discourse \\
$[0.08,\;0.15,\;0.77]$ & 14228 & Multi word named entity detector (proper nouns, locations etc.) \\
$[0.00,\;0.00,\;1.00]$ & 15027 & Activates on last token of capitalized names (person, location etc.) \\
$[0.00,\;0.18,\;0.82]$ & 6746 & Detects deverbal nouns / nominalizations, abstract/eventive nouns formed from verbs \\
$[0.00,\;0.10,\;0.90]$ & 5317 & - \\
$[0.01,\;0.23,\;0.76]$ & 14629 & Newline detector \\
$[0.10,\;0.08,\;0.82]$ & 13117 & Newline detector \\
$[0.00,\;0.00,\;1.00]$ & 15129 & First name detector \\
$[0.00,\;0.01,\;0.99]$ & 14623 & Detects prepositions \\
\bottomrule
\end{tabular}
}%
\caption{\textbf{3-way L1‑Sparsity Crosscoder Annotation for Pythia-1B | Comparison 1B \compar{} 4B \compar{} 286B.} \relie{} shows 3-way one-versus-all \relie{} vector; Interpreted Function provides a description if a linguistic role was detected, and ``--'' otherwise. Rows are grouped by checkpoint specificity according to \relie{}: features dominated by one checkpoint (1B, 4B, 286B specific); pairwise shared features (1B–4B, 1B–286B, 4B–286B shared); and shared across all (1B–4B–286B shared). A missing group means no such features found in the top-10 IE features of all checkpoints. \relie{}-based triplet comparisons reveal that earlier checkpoints (\eg{}, 1B and 4B) primarily detect low-level lexical and morphological patterns such as suffixes and irregular plurals, whereas later checkpoints (\eg{}, 286B) increasingly specialize in higher-level syntactic and semantic functions, including named entity, nominalization, and technical discourse related noun detection.}
\label{tab:3way_annotation_pythia}
\end{table*}

\begin{table*}[!ht]
\centering
\resizebox{0.95\linewidth}{!}{%
\begin{tabular}{crl}
\toprule
\mythead{RelIE} & \mythead{FeatID} & \mythead{Interpreted Function} \\
\midrule
\multicolumn{3}{l}{\bfseries 4B specific} \\
\midrule
$[1.00,\;0.00,\;0.00]$ & 675 & Regular plural noun detector, activates on  final tokens of regular plural nouns and promotes new word completions \\
$[1.00,\;0.00,\;0.00]$ & 10707 & - \\
$[0.99,\;0.00,\;0.01]$ & 8433 & - \\
$[1.00,\;0.00,\;0.00]$ & 15961 & Nominalization feature that detects deverbal and derivational nouns (\ie{}, \emph{-ance}, \emph{-ion}, \emph{-ing} etc.) \\
$[0.64,\;0.24,\;0.12]$ & 3269 & Singular noun detector preceeded by \emph{This}, promotes singular verb conjugations \\

\multicolumn{3}{l}{\bfseries 4B-33B shared} \\
\midrule
$[0.36,\;0.34,\;0.30]$ & 702 & Plural noun detector preceeded by plural quantifier (\eg{} \emph{most}, \emph{some}), promotes plural verb conjugation \\
\multicolumn{3}{l}{\bfseries 4B-3048B shared} \\
\midrule
$[0.38,\;0.25,\;0.37]$ & 16117 & Stock‐ticker/exchange‐code detector \\
\multicolumn{3}{l}{\bfseries 33B specific} \\
\midrule
$[0.00,\;1.00,\;0.00]$ & 5966 & Detects commas followed by parenthetical clauses \\
$[0.03,\;0.90,\;0.07]$ & 7527 & Headline/title-case text detector \\
$[0.00,\;0.80,\;0.19]$ & 10924 & Detects first names that aren't followed up by last names \\
$[0.00,\;1.00,\;0.00]$ & 10692 & Regular plural noun detector \\
\multicolumn{3}{l}{\bfseries 33B-3048B shared} \\
\midrule
$[0.00,\;0.50,\;0.50]$ & 9908 & Noun/head-of-NP detector (both common and proper, singular and plural, simple or compound) \\
$[0.00,\;0.50,\;0.50]$ & 15717 & Plural noun detector for plural people nouns highlighting attributes or jobs \\
$[0.00,\;0.52,\;0.48]$ & 14569 & Detects last token of multi-token first names followed by last names \\
$[0.00,\;0.46,\;0.54]$ & 9230 & \emph{-s/-es} noun inflection detector on stems that could have been verbs but become nouns \\
$[0.02,\;0.65,\;0.33]$ & 847 & Detects final token of first names to be followed by last names \\
\multicolumn{3}{l}{\bfseries 3048B specific} \\
\midrule
$[0.25,\;0.24,\;0.51]$ & 3515 & Newline detector \\
$[0.00,\;0.29,\;0.71]$ & 13176 & Detects plural countable objects \\
$[0.01,\;0.24,\;0.75]$ & 8084 & - \\
$[0.09,\;0.06,\;0.85]$ & 1469 & - \\
$[0.00,\;0.00,\;1.00]$ & 1656 & Detects punctuation or conjunction preceeded by named entities, promotes certain verb conjugations \\
$[0.00,\;0.00,\;1.00]$ & 6319 & Newline detector \\
$[0.00,\;0.00,\;1.00]$ & 5550 & Newline detector that promotes certain sentence beginnings \\
\bottomrule
\end{tabular}
}%
\caption{\textbf{3-way L1‑Sparsity Crosscoder Annotation for OLMo-1B | Comparison 4B \compar{} 33B \compar{} 3048B.} Similar to Pythia, OLMo progresses from detecting lower-level lexical and morphological patterns in early checkpoints to more abstract grammatical and noun-phrase features later on, but OLMo may be retaining a stronger persistence of surface-level detectors (\eg{}, newlines, suffixes) compared to Pythia’s sharper shift.}
\label{tab:3way_annotation_olmo}
\end{table*}
\begin{table*}[!ht]
\centering
\resizebox{1.0\linewidth}{!}{%
\begin{tabular}{crll}
\toprule
\mythead{RelIE} & \mythead{FeatID} & \mythead{Interpreted Function} & \mythead{Languages} \\
\midrule
\multicolumn{4}{l}{\bfseries 6B specific} \\
\midrule
$[1.00,\;0.00,\;0.00]$ & 3672 & Detects ellipsis and question/exclamation marks & arb,eng,fra,hin,por,spa \\
$[0.78,\;0.14,\;0.07]$ & 7122 & Main-verb head detector & eng \\
$[0.83,\;0.14,\;0.03]$ & 10388 & Plural noun detector for several languages (\eg{}, \emph{players}, \emph{usários} - users, \emph{al naas} - the people, {\dn mEhlAe{\qva}} - ladies) & arb,eng,fra,hin,por,spa \\
$[0.72,\;0.23,\;0.06]$ & 9163 & Noun‑phrase head detector of multi‑word noun chunk, activates on the key content (noun or adjective) that carries the meaning & eng,fra,por,spa \\
$[0.62,\;0.17,\;0.20]$ & 15288 & Detects \emph{Ev}, \emph{ev}, \emph{év} subtokens in different languages & eng,fra,por,spa \\
$[0.59,\;0.17,\;0.24]$ & 5704 & - & - \\
\multicolumn{4}{l}{\bfseries 6B-55B shared} \\
\midrule
$[0.35,\;0.41,\;0.24]$ & 15248 & Detects the token \emph{that} only in English & eng \\
$[0.36,\;0.36,\;0.28]$ & 12525 & Boss concept detector (\eg{}, \emph{chief}, \emph{jefe}, \emph{chefs}, \emph{chefe}, {\dn \7mHy} - boss) & eng,fra,hin,por,spa \\
$[0.56,\;0.27,\;0.17]$ & 15758 & Detects head of multi-token or compound nouns & eng \\
\multicolumn{4}{l}{\bfseries 55B specific} \\
\midrule
$[0.15,\;0.63,\;0.22]$ & 10862 & Predicts tokens related to the concept or form of \emph{consult} (\eg{}, \zh{咨询}, \emph{fasr} - explain) & eng,fra,por,spa \\
\multicolumn{4}{l}{\bfseries 55B-341B shared} \\
\midrule
$[0.05,\;0.31,\;0.64]$ & 6997 & Proper‑noun/ID detector that activates on named-entity heads & arb,eng,fra,hin,por,spa \\
\multicolumn{4}{l}{\bfseries 341B specific} \\
\midrule
$[0.00,\;0.00,\;1.00]$ & 15193 & Detects punctuation and parenthesis & arb,eng,fra,hin,spa,zh \\
$[0.00,\;0.00,\;1.00]$ & 2598 & - & - \\
$[0.00,\;0.00,\;1.00]$ & 12151 & - & - \\
$[0.00,\;0.00,\;1.00]$ & 9110 & - & - \\
$[0.00,\;0.00,\;1.00]$ & 7066 & - & - \\
$[0.00,\;0.00,\;1.00]$ & 6461 & - & - \\
\multicolumn{4}{l}{\bfseries 6B-55B-341B shared} \\
\midrule
$[0.35,\;0.32,\;0.32]$ & 12140 & Multilingual relative pronoun detector (\eg{}, \emph{que}, \emph{that}, \emph{who}, \emph{aladhi}) & arb,eng,fra,por,spa \\
$[0.32,\;0.31,\;0.37]$ & 4610 & Phrasal‑verb/PP‑complement detector that fires on the first token of a verb‑plus‑particle or adjective‑plus‑preposition pattern & eng,fra,por,spa \\
$[0.39,\;0.26,\;0.35]$ & 5819 & Activates most on new beginning of clauses right after a punctuation and wanes until a new clause & arb,eng,fra,spa \\
\bottomrule
\end{tabular}
}%
\caption{\textbf{3-way L1‑Sparsity Crosscoder CLAMS French/English Annotation for BLOOM-1B  | 6B \compar{} 55B \compar{} 341B.} The languages column depicts which languages appeared to also use this feature when observing the feature's top-activating sentences. Early checkpoints often rely on language-specific patterns (\eg{}, English token detectors, suffix patterns), but later checkpoints increasingly learn cross-lingual and language-shared features, such as multilingual pronoun, noun-phrase, and clause detectors.}
\label{tab:3way_annotation_bloom_clams_fraeng}
\end{table*}

\begin{table*}[!ht]
\centering
\resizebox{1.0\linewidth}{!}{%
\begin{tabular}{crll}
\toprule
\mythead{RelIE} & \mythead{FeatID} & \mythead{Interpreted Function} & \mythead{Languages} \\
\midrule
\multicolumn{4}{l}{\bfseries 6B specific} \\
\midrule
$[0.94,\;0.02,\;0.04]$ & 3672 & Detects ellipsis and question/exclamation marks & arb,eng,fra,hin,por,spa \\
$[1.00,\;0.00,\;0.00]$ & 5273 & Detects commas and full stop & eng,fra,hin,por,spa \\
$[0.45,\;0.28,\;0.26]$ & 849 & Plural nouns / noun compounds that typically are the subject at BOS, promotes plural verb completion & eng \\
$[0.50,\;0.24,\;0.26]$ & 10974 & 3rd person plural pronoun \emph{they} detector & eng \\
$[0.59,\;0.24,\;0.17]$ & 9163 & Noun‑phrase head detector, activates on the key content word (noun or adjective) that carries the meaning of multi‑word noun chunk & eng,fra,por,spa \\
$[0.70,\;0.23,\;0.07]$ & 15758 & Detects head of multi token or compound nouns & eng \\
\multicolumn{4}{l}{\bfseries 6B-55B shared} \\
\midrule
$[0.44,\;0.31,\;0.25]$ & 10235 & Adverbial connectives (\eg{}, \emph{then}, \emph{tambien}, \emph{also}, \emph{tambem}) & eng,por,spa \\
$[0.40,\;0.35,\;0.26]$ & 4332 & Detects regular plural nouns & eng \\
$[0.49,\;0.37,\;0.15]$ & 5819 & Activates most on new beginning of clauses right after a punctuation and wanes until a new clause & arb,eng,fra,spa \\
\multicolumn{4}{l}{\bfseries 6B-341B shared} \\
\midrule
$[0.40,\;0.21,\;0.39]$ & 5704 & - & - \\
$[0.45,\;0.23,\;0.32]$ & 7007 & People-related regular plural nouns (\eg{}, \emph{workers}, \emph{investors}, \emph{experts}) & eng \\
\multicolumn{4}{l}{\bfseries 55B specific} \\
\midrule
$[0.19,\;0.52,\;0.29]$ & 14073 & Detects noun beginnings related to academic write‑ups (\eg{}, \emph{dissertation}, \emph{thesis}, {\dn Enb{\qva}D} - Hindi, \emph{essai} - French) & eng,fra,hin \\
$[0.00,\;1.00,\;0.00]$ & 8086 & Detects repeated interpuncts, most likely a training data artifact & - \\
\multicolumn{4}{l}{\bfseries 341B specific} \\
\midrule
$[0.00,\;0.00,\;1.00]$ & 7694 & - & - \\
$[0.00,\;0.00,\;1.00]$ & 9110 & - & - \\
$[0.00,\;0.00,\;1.00]$ & 15193 & Detects punctuation and parenthesis & arb,eng,fra,hin,spa,zh \\
$[0.10,\;0.30,\;0.60]$ & 11280 & - & - \\
$[0.00,\;0.00,\;1.00]$ & 6461 & - & - \\
$[0.01,\;0.05,\;0.93]$ & 14020 & - & - \\
$[0.00,\;0.04,\;0.96]$ & 2598 & - & - \\
$[0.00,\;0.00,\;1.00]$ & 1638 & Detects brackets/parenthesis/punctuation & eng,fra,por,spa \\
$[0.00,\;0.00,\;1.00]$ & 7066 & - & - \\
$[0.00,\;0.00,\;1.00]$ & 12151 & - & - \\
\bottomrule
\end{tabular}
}%
\caption{\textbf{3-way L1‑Sparsity Crosscoder MultiBLiMP English Annotation for BLOOM-1B  | 6B \compar{} 55B \compar{} 341B.} While this annotation focuses on finding features with just English examples, many features in BLOOM still activate across multiple languages due to cross-lingual representations, showing that the model often leverages multilingual patterns such as shared connectives, pronouns, and beginning of clause detectors.}
\label{tab:3way_annotation_bloom_multiblimp_eng}
\end{table*}

\begin{table*}[!ht]
\centering
\resizebox{1.0\linewidth}{!}{%
\begin{tabular}{crll}
\toprule
\mythead{RelIE} & \mythead{FeatID} & \mythead{Interpreted Function} & \mythead{Languages} \\
\midrule
\multicolumn{4}{l}{\bfseries 6B specific} \\
\midrule
$[1.00,\;0.00,\;0.00]$ & 16271 & Detects comma, punctuation possibly to mark new clauses & eng,fra,hin,por,spa,zh \\
$[0.91,\;0.01,\;0.08]$ & 3672 & Detects ellipsis and question/exclamation marks & arb,eng,fra,hin,por,spa \\
$[1.00,\;0.00,\;0.00]$ & 5273 & Detects commas and full stop & eng,fra,hin,por,spa \\
$[1.00,\;0.00,\;0.00]$ & 7529 & Detects comma, punctuation possibly to mark new clauses & eng,fra,hin,por,spa,zh \\
$[1.00,\;0.00,\;0.00]$ & 11289 & Detects comma, punctuation possibly to mark new clauses & eng,fra,hin,por,spa,zh \\
$[0.51,\;0.23,\;0.26]$ & 13809 & 1st person singular pronoun detector in diff surface forms (\eg{}, \emph{j'}, \emph{Je}, \emph{je}, \emph{yo}) & fra,spa \\
$[0.63,\;0.24,\;0.12]$ & 8643 & Plural first person pronoun detector, multilingual & arb,fra,por,spa \\
\multicolumn{4}{l}{\bfseries 6B-55B shared} \\
\midrule
$[0.63,\;0.26,\;0.12]$ & 14472 & Plural second person pronoun detector & fra \\
\multicolumn{4}{l}{\bfseries 6B-341B shared} \\
\midrule
$[0.33,\;0.25,\;0.42]$ & 5819 & Activates most on new beginning of clauses right after a punctuation and wanes until a new clause & arb,eng,fra,spa \\
$[0.45,\;0.24,\;0.31]$ & 12522 & capitalized \emph{Je} surface form detector, 1st person singular in French but also activates on names and \emph{Jeux} game and \emph{Yo} & eng,fra,spa \\
\multicolumn{4}{l}{\bfseries 55B specific} \\
\midrule
$[0.00,\;1.00,\;0.00]$ & 14378 & - & - \\
$[0.00,\;1.00,\;0.00]$ & 6506 & - & - \\
$[0.00,\;1.00,\;0.00]$ & 13738 & - & - \\
$[0.29,\;0.61,\;0.10]$ & 16196 & - & - \\
$[0.20,\;0.60,\;0.21]$ & 2471 & Detects plural second person pronouns, multilingual & arb,eng,fra,hin \\
\multicolumn{4}{l}{\bfseries 341B specific} \\
\midrule
$[0.00,\;0.00,\;1.00]$ & 15193 & Detects punctuation and parenthesis & arb,eng,fra,hin,spa,zh \\
$[0.09,\;0.01,\;0.90]$ & 14020 & - & - \\
$[0.02,\;0.00,\;0.98]$ & 12151 & - & - \\
$[0.00,\;0.00,\;1.00]$ & 1638 & Detects brackets/parenthesis/punctuation & eng,fra,por,spa \\
$[0.03,\;0.00,\;0.97]$ & 7694 & - & - \\
$[0.02,\;0.01,\;0.97]$ & 7066 & - & - \\
$[0.02,\;0.00,\;0.98]$ & 6461 & - & - \\
$[0.00,\;0.00,\;1.00]$ & 2598 & - & - \\
$[0.03,\;0.00,\;0.97]$ & 9110 & - & - \\
\bottomrule
\end{tabular}
}%
\caption{\textbf{3-way L1‑Sparsity Crosscoder MultiBLiMP French Annotation for BLOOM-1B  | 6B \compar{} 55B \compar{} 341B.} Similar to the English examples only analysis, we find that many features found with a French dataset activate across multiple languages, reflecting BLOOM’s shared cross-lingual representations. Some early detectors remain French-focused (\eg{}, surface-form capitalization and plural pronouns), while later features, like punctuation and pronoun detectors, consolidate to be multilingual.}
\label{tab:3way_annotation_bloom_multiblimp_fra}
\end{table*}

\begin{table*}[!ht]
\centering
\resizebox{1.0\linewidth}{!}{%
\begin{tabular}{crll}
\toprule
\mythead{RelIE} & \mythead{FeatID} & \mythead{Interpreted Function} & \mythead{Languages} \\
\midrule
\multicolumn{4}{l}{\bfseries 6B specific} \\
\midrule
$[0.79,\;0.00,\;0.21]$ & 14483 & Locative/in‑marker {\dn m\?{\qva}} & hin \\
$[0.67,\;0.18,\;0.16]$ & 4192 & Present‑habitual marker using {\dn jAnA} / {\dn honA} for singular‑masculine subjects & hin \\
$[0.67,\;0.22,\;0.11]$ & 1174 & The -{\dn n\?} participle ending marking habitual aspect, here with a masculine‑plural (respectful) participle & hin \\
$[0.66,\;0.22,\;0.13]$ & 8471 & Subordinating conjunction {\dn Ek} (``that'') introducing subordinate clauses & hin \\
$[0.51,\;0.24,\;0.25]$ & 2539 & Perfective participle plural (and oblique‑singular) ending used adjectivally & hin \\
\multicolumn{4}{l}{\bfseries 6B-55B shared} \\
\midrule
$[0.45,\;0.36,\;0.20]$ & 6215 & Detects abstract singular nouns (chance, permission, responsibility, order, signal, shelter, danger, etc.) & hin \\
$[0.40,\;0.54,\;0.07]$ & 4338 & Light‑verb root {\dn kr} used as ``do'' auxiliary in compound verbs (root + another light‑verb + conjugation) & hin \\
\multicolumn{4}{l}{\bfseries 6B-341B shared} \\
\midrule
$[0.64,\;0.09,\;0.27]$ & 3969 & Feminine possessive marker {\dn kF} (``of''/’s) & hin \\
$[0.68,\;0.05,\;0.27]$ & 4361 & Masculine possessive marker {\dn kA} (``of''/’s) & hin \\
\multicolumn{4}{l}{\bfseries 55B specific} \\
\midrule
$[0.19,\;0.61,\;0.20]$ & 11884 & Marker detecting the second noun or second element in a compound & hin \\
\multicolumn{4}{l}{\bfseries 55B-341B shared} \\
\midrule
$[0.26,\;0.31,\;0.42]$ & 4579 & Nominalizer of ``to be,'' functioning as a gerund (``its being X,'' ``because of X being'') & hin \\
$[0.05,\;0.43,\;0.52]$ & 643 & Detects first token of verbs / verb roots that appear before subject number conjugation & hin \\
$[0.04,\;0.37,\;0.58]$ & 2526 & Perfective aspect marker in compounds like {\dn EkyA gyA} (``did/gave'' in the perfective) & hin \\
$[0.02,\;0.54,\;0.44]$ & 1082 & Inflection of the verb ``to be'' ({\dn ho}) in the subjunctive/continuous mood (\eg{}, {\dn ho sktA, ho \7ckA}) indicating possibility or completed action & hin \\
\multicolumn{4}{l}{\bfseries 341B specific} \\
\midrule
$[0.20,\;0.21,\;0.59]$ & 11856 & Negation marker {\dn nhF{\qva}} placed before verbs & hin \\
$[0.00,\;0.00,\;1.00]$ & 2598 & - & - \\
$[0.00,\;0.00,\;1.00]$ & 14020 & - & - \\
\multicolumn{4}{l}{\bfseries 6B-55B-341B shared} \\
\midrule
$[0.48,\;0.26,\;0.26]$ & 2563 & Plural pronoun marker for ``you/you all'' or ``us/we'' & hin \\
\bottomrule
\end{tabular}
}%
\caption{\textbf{3-way L1‑Sparsity Crosscoder MultiBLiMP Hindi Annotation for BLOOM-1B  | 6B \compar{} 55B \compar{} 341B.} When using just Hindi subject--verb agreement examples to find highest IE features, most features remain strongly language-specific, focusing on tense, aspect, case, and possessive markers, with fewer cross-lingual activations compared to English and French. This may be due to Hindi’s lower representation in the BLOOM training data (2\%) relative to French (15\%), which limits the emergence of more shared, language-agnostic detectors.}
\label{tab:3way_annotation_bloom_multiblimp_hin}
\end{table*}

\section{Additional Analyses}
\label{sec:appendix_analyses}

\begin{figure*}[t]
    \centering
    \includegraphics[width=0.9\linewidth]{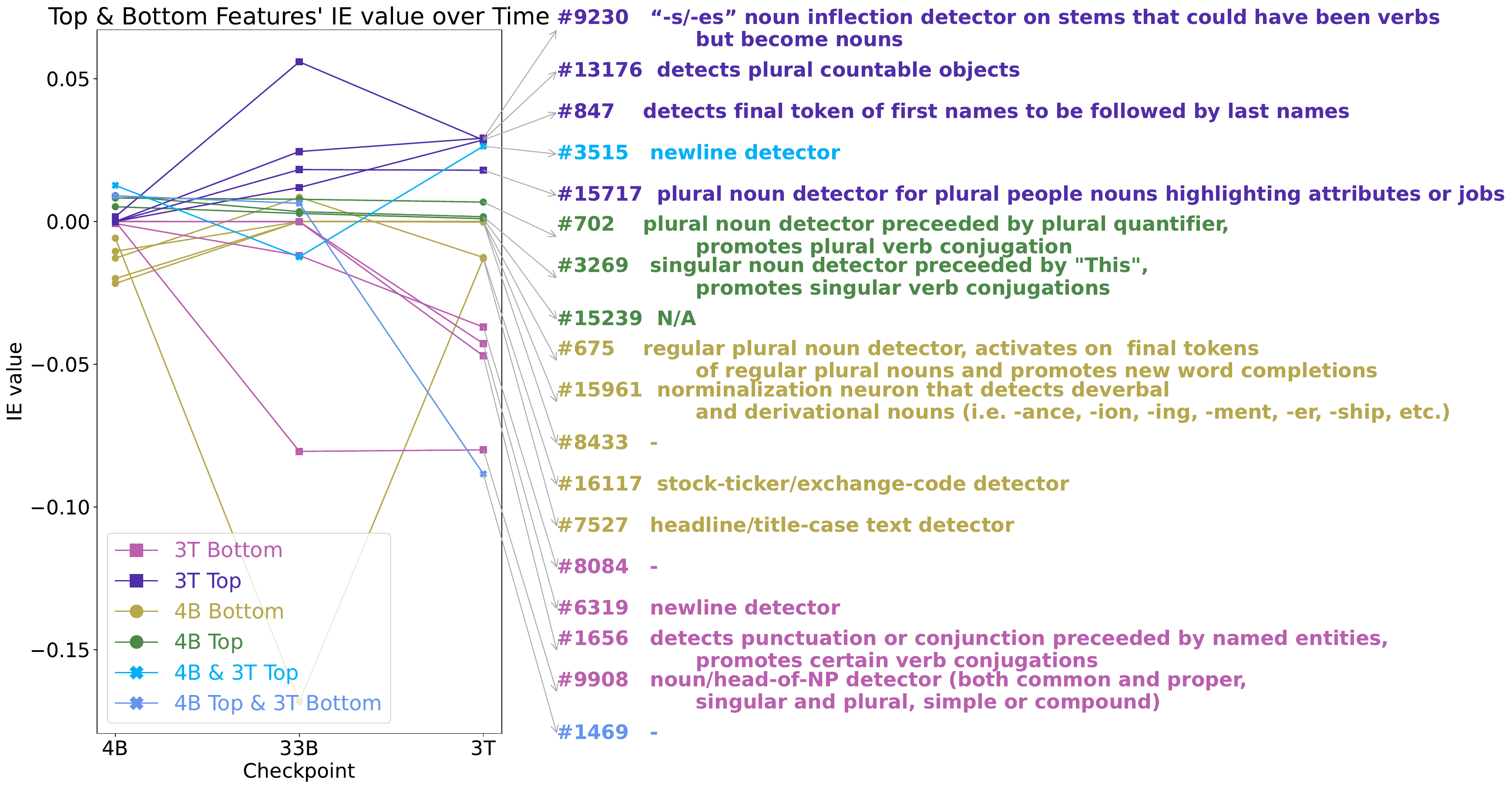}
    \caption{\textbf{IE evolution of Top-5 \& Bottom-5 Features for OLMo-1B checkpoints 4B \& 3T.} IEs are calculated using BLiMP subject--verb agreement tasks. ``--'' means the feature was not interpretable. In some cases, a feature can belong to multiple categories at once. Some low-level features, such as newline detectors, persist across training, whereas the usage of simpler lexical detectors fade as more abstract grammatical pattern detectors emerge.}
    \label{fig:ie-evolution-olmo}
\end{figure*}

In Fig.~\ref{fig:ie-evolution-olmo}, we provide an additional plot showing the IE evolution of OLMo’s top and bottom five IE features at the 4B and 3T checkpoints, complementing the Pythia evolution figure in the main paper (Fig.~\ref{fig:ie-evolution-pythia}).

\begin{figure*}[th]
    \centering
    \includegraphics[width=1.0\textwidth]{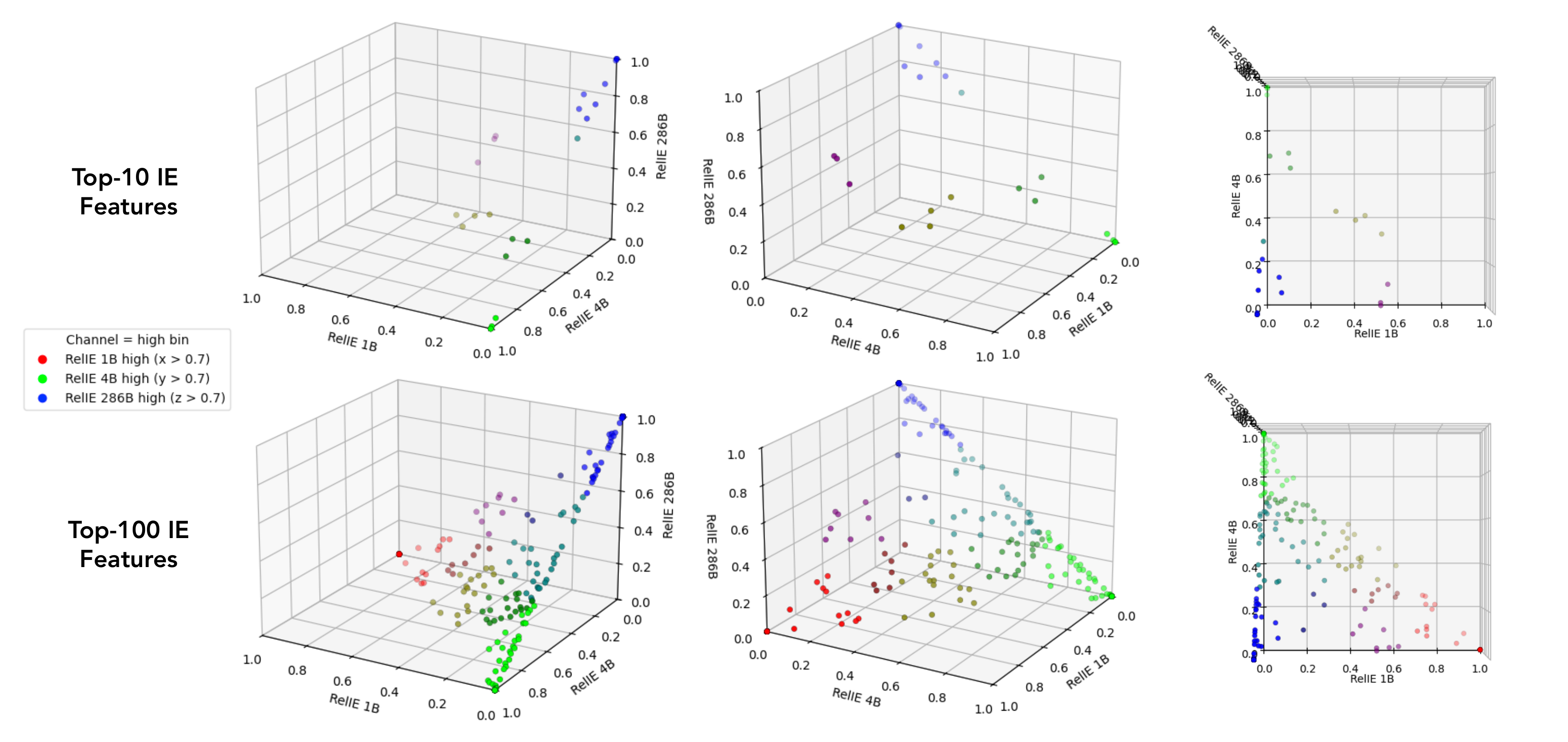}
    \caption{\textbf{\relie{} of Top-10 and 100 IE features on BLiMP for Pythia-1B checkpoints \{1B, 4B, 286B\}.} 
    Distinct clusters near each corner indicate checkpoint-specific features. In the Top-10 row, the 4B-286B pair and features shared across all three checkpoints dominate, whereas the 1B-286B pair have relatively fewer shared features. Additionally, the checkpoint-specific regions for 4B and 286B are noticeably denser, suggesting a richer set of unique features in these models that are trained on more data than 1B.}
    \label{fig:monolingual-overlap-full}
\end{figure*}

We also include an additional monolingual overlap plot in Fig.~\ref{fig:monolingual-overlap-full}, which shows the top 10 IE features in addition to the top 100 shown in the main paper (Fig.~\ref{fig:monolingual-overlap}).

\begin{figure*}[th]
    \centering
    \begin{subfigure}{0.32\linewidth}
      \centering
      \includegraphics[width=\textwidth]{figs/multilingual_overlap_3way/multilang_overlap_topk10_SV-num_0.png}
    \end{subfigure}
    \begin{subfigure}{0.32\linewidth}
      \centering
      \includegraphics[width=\textwidth]{figs/multilingual_overlap_3way/multilang_overlap_topk10_SV-num_1.png}
    \end{subfigure}
    \begin{subfigure}{0.32\linewidth}
      \centering
      \includegraphics[width=\textwidth]{figs/multilingual_overlap_3way/multilang_overlap_topk10_SV-num_2.png}
    \end{subfigure}
    \textbf{(a) Feature set overlap --- SV-\# (6B, 55B, 341B)}
    \vspace{1ex}
    
    \begin{subfigure}{0.32\linewidth}
      \centering
      \includegraphics[width=\textwidth]{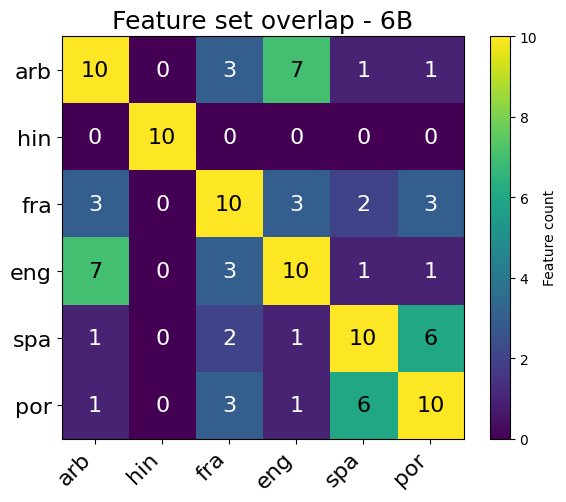}
    \end{subfigure}
    \begin{subfigure}{0.32\linewidth}
      \centering
      \includegraphics[width=\textwidth]{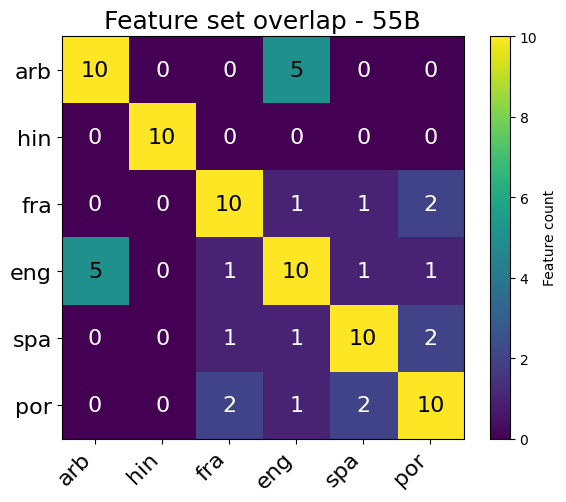}
    \end{subfigure}
    \begin{subfigure}{0.32\linewidth}
      \centering
      \includegraphics[width=\textwidth]{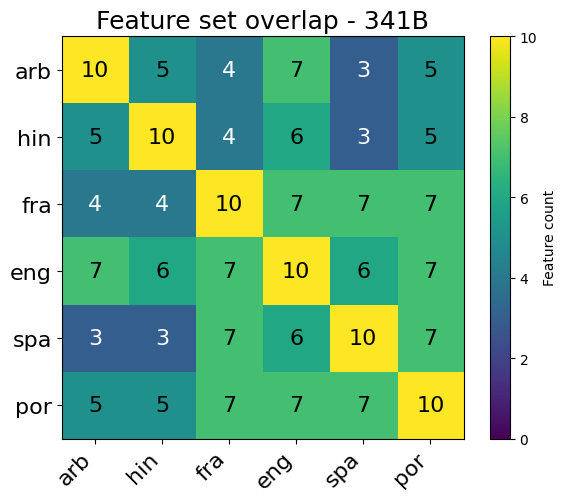}
    \end{subfigure}
    \textbf{(b) Feature set overlap --- SV-P (6B, 55B, 341B)}
    \vspace{1ex}
    
    \begin{subfigure}{0.32\linewidth}
      \centering
      \includegraphics[width=\textwidth]{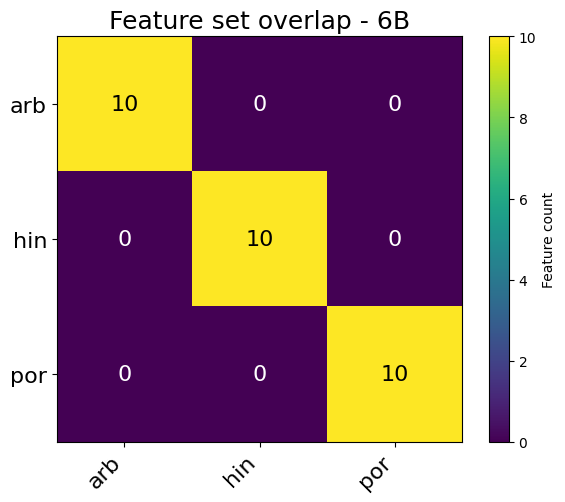}
    \end{subfigure}
    \begin{subfigure}{0.32\linewidth}
      \centering
      \includegraphics[width=\textwidth]{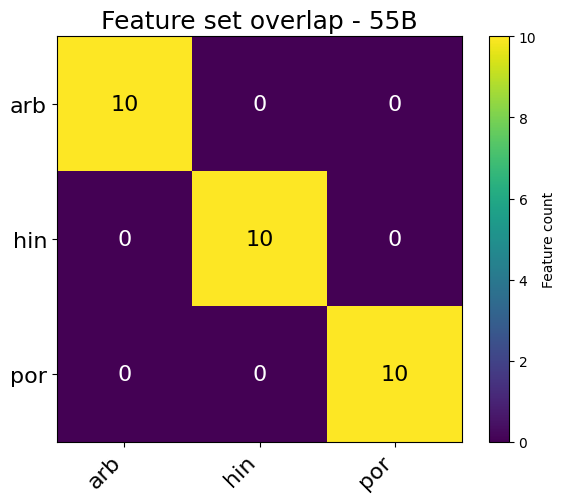}
    \end{subfigure}
    \begin{subfigure}{0.32\linewidth}
      \centering
      \includegraphics[width=\textwidth]{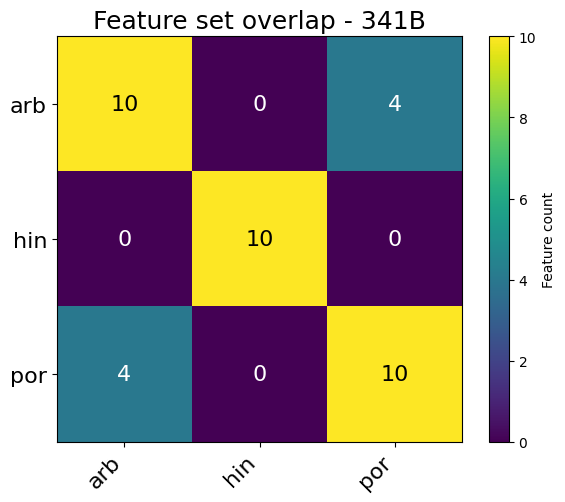}
    \end{subfigure}
    \textbf{(c) Feature set overlap --- SV-G (6B, 55B, 341B)}
    \vspace{1ex}
    \caption{\textbf{Top-10 IE feature overlap in BLOOM-1B across languages and subtasks with the 3-way comparison.} Across BLOOM-1B checkpoints, feature overlap is generally higher among script-sharing languages (\eg{}, English, French, Spanish, Portuguese) and increases across pretraining, while languages like Arabic and Hindi, which are less frequent in the training data and use different scripts, show relatively less cross-lingual feature sharing.}
    \label{fig:multilingual-overlap-3way}
\end{figure*}

\begin{figure*}[th]
    \centering
    \begin{subfigure}{0.32\linewidth}
      \centering
      \includegraphics[width=\textwidth]{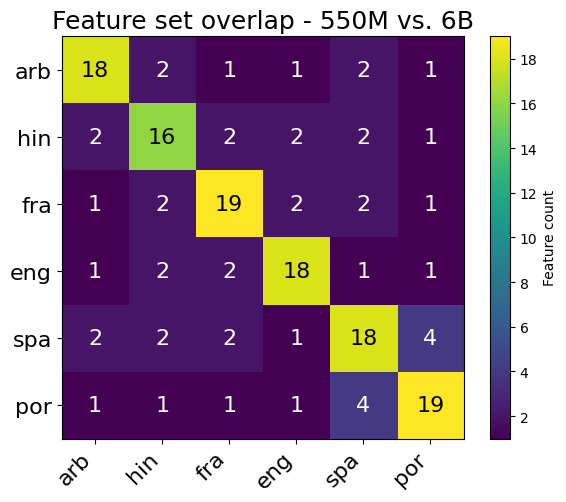}
    \end{subfigure}
    \begin{subfigure}{0.32\linewidth}
      \centering
      \includegraphics[width=\textwidth]{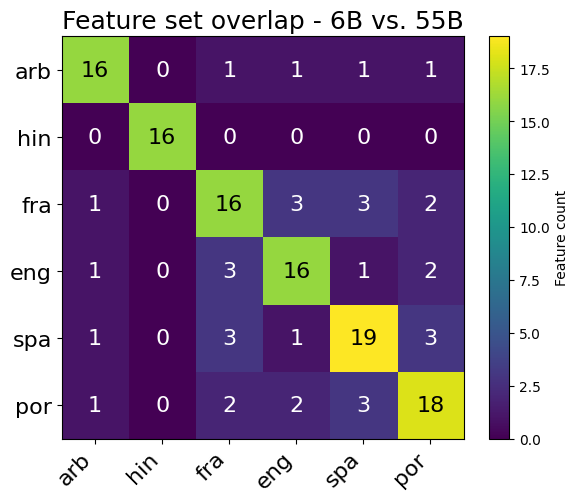}
    \end{subfigure}
    \begin{subfigure}{0.32\linewidth}
      \centering
      \includegraphics[width=\textwidth]{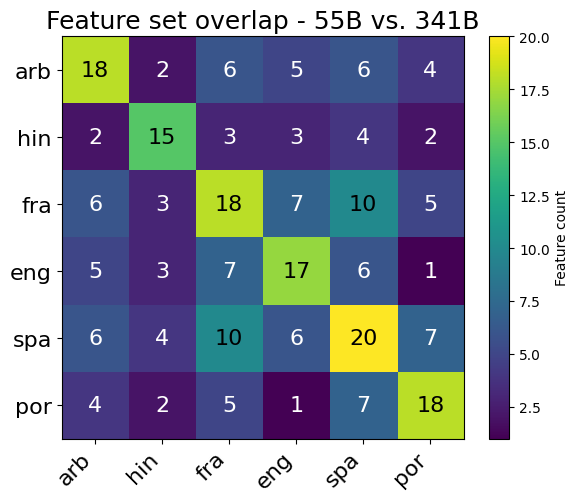}
    \end{subfigure}
    \textbf{(a) Feature set overlap --- SV-\# (550M vs. 6B, 6B vs. 55B, 55B vs. 341B)}
    \vspace{1ex}
    
    \begin{subfigure}{0.32\linewidth}
      \centering
      \includegraphics[width=\textwidth]{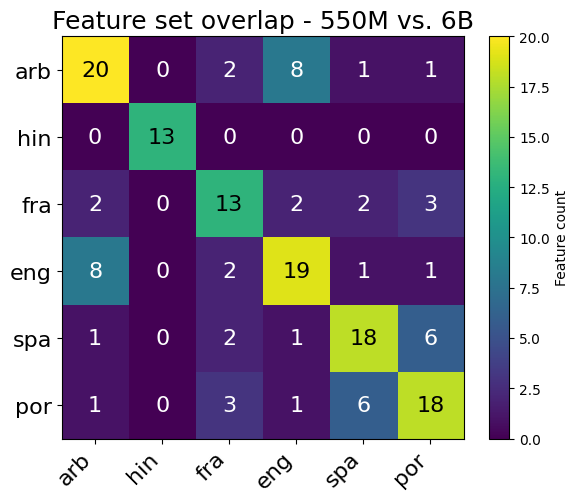}
    \end{subfigure}
    \begin{subfigure}{0.32\linewidth}
      \centering
      \includegraphics[width=\textwidth]{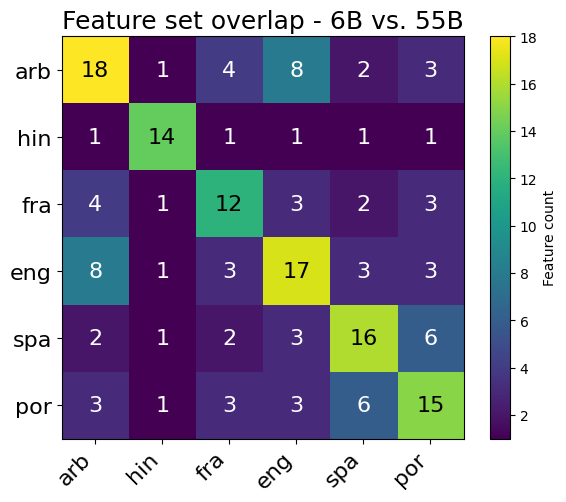}
    \end{subfigure}
    \begin{subfigure}{0.32\linewidth}
      \centering
      \includegraphics[width=\textwidth]{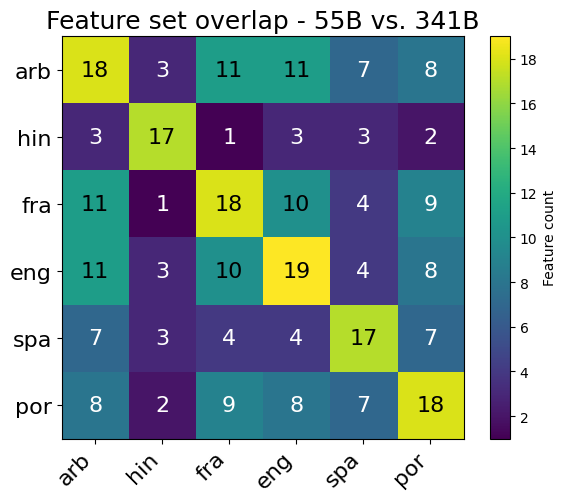}
    \end{subfigure}
    \textbf{(b) Feature set overlap --- SV-P (550M vs. 6B, 6B vs. 55B, 55B vs. 341B)}
    \vspace{1ex}
    
    \begin{subfigure}{0.32\linewidth}
      \centering
      \includegraphics[width=\textwidth]{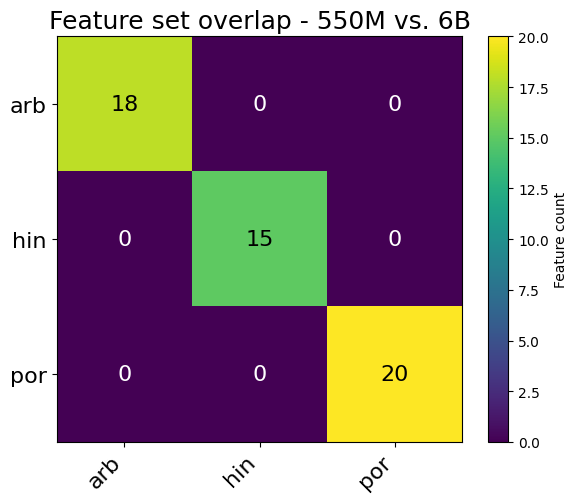}
    \end{subfigure}
    \begin{subfigure}{0.32\linewidth}
      \centering
      \includegraphics[width=\textwidth]{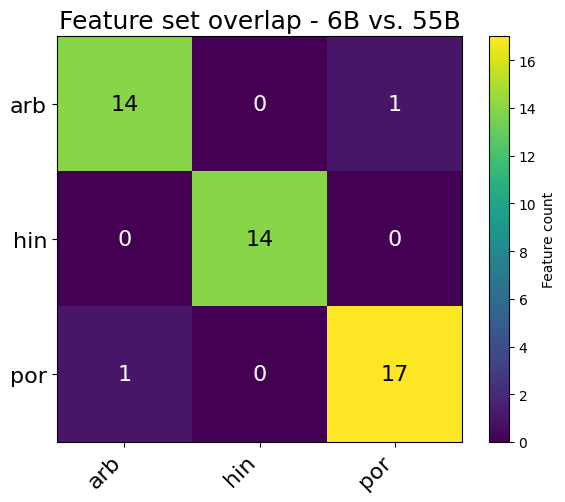}
    \end{subfigure}
    \begin{subfigure}{0.32\linewidth}
      \centering
      \includegraphics[width=\textwidth]{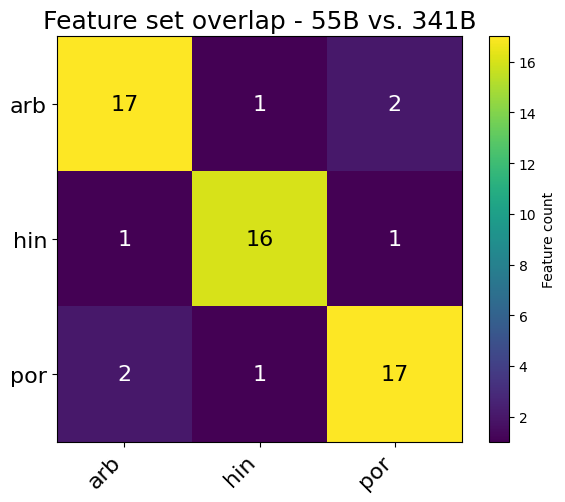}
    \end{subfigure}
    \textbf{(c) Feature set overlap --- SV-G (550M vs. 6B, 6B vs. 55B, 55B vs. 341B)}
    \vspace{1ex}
    \caption{\textbf{Top-10 IE feature overlap (per checkpoint) in BLOOM-1B across languages and subtasks with the 2-way comparison.} The 2-way comparison shows a similar pattern to the 3-way analysis, with high feature overlap among related languages (\eg{}, English, French, Spanish, Portuguese). Notably, in the 55B vs.\ 341B comparison, Arabic---despite not being Indo-European---shares more features than Hindi, suggesting better cross-lingual generalization for Arabic at later checkpoints. One explanation can be that Arabic is more prevalent in the training data (5\%) than Hindi (2\%).}
    \label{fig:multilingual-overlap-2way}
\end{figure*}

Finally, we present the top-10 significant feature overlap counts from the multilingual two-way and three-way comparisons for all MultiBLiMP subtasks (number, person, and gender agreement) in Figures~\ref{fig:multilingual-overlap-3way} and \ref{fig:multilingual-overlap-2way}, respectively.

\end{document}